\documentclass[10pt,twocolumn,letterpaper]{article}
\usepackage{iccv}
\usepackage{times} 
\usepackage{graphicx}
\usepackage{amsmath}
\usepackage{amssymb}
\usepackage{picture}
\usepackage{lipsum}
\usepackage[%
pagebackref=true,
breaklinks=true,
letterpaper=true,
colorlinks,
bookmarks=false]{hyperref}
\usepackage{cleveref}
\usepackage{multirow}
\usepackage{diagbox}
\usepackage{booktabs}
\usepackage{pbox}

\crefname{section}{sec.}{sec.}
\Crefname{section}{Sec.}{Sec.}

\iccvfinalcopy 
\ificcvfinal\fi

\newcommand{\myparagraph}[1]{\vspace{0.1cm}\noindent\textbf{#1.}}
\newcommand{\define}{\ensuremath{\overset{\cdot}{=}}}

\newcommand{\netname}{VpDR-Net\xspace}
\newcommand{\netnamedepth}{BerHu-Net\xspace}
\newcommand{\PhiVP}{\ensuremath{\Phi_\text{vp}}}
\newcommand{\PhiPCL}{\ensuremath{\Phi_\text{pcl}}}
\newcommand{\PhiDepth}{\ensuremath{\Phi_\text{depth}}}
\newcommand{\lossPCL}{\ensuremath{\ell_\text{pcl}}}
\newcommand{\tb}[1]{\textbf{#1}\xspace}

\title{Learning 3D Object Categories by Looking Around Them}

\author{
David Novotny$^{1,2}$ ~ ~ Diane Larlus$^2$ ~ ~ Andrea Vedaldi$^1$ \\
\centering
\begin{minipage}{.4\textwidth}
\centering
$^1$\small{Visual Geometry Group\\Dept. of Engineering Science, University of Oxford\\}
{\tt\small \{david,vedaldi\}@robots.ox.ac.uk} 
\end{minipage} 
\begin{minipage}{.4\textwidth}
\centering
$^2$\small{Computer Vision Group\\Naver Labs Europe\\}
{\tt\small diane.larlus@naverlabs.com} 
\end{minipage}
}

\begin{document} 

\maketitle

\begin{abstract} 
Traditional approaches for learning 3D object categories use either synthetic data or manual supervision. In this paper, we propose a method which does not require manual annotations and is instead cued by observing objects from a moving vantage point. Our system builds on two innovations: a Siamese viewpoint factorization network that robustly aligns different videos together without explicitly comparing 3D shapes; and a 3D shape completion network that can extract the full shape of an object from partial observations. We also demonstrate the benefits of configuring networks to perform \emph{probabilistic predictions} as well as of \emph{geometry-aware data augmentation} schemes. We obtain state-of-the-art results on publicly-available benchmarks. 
\end{abstract}

\section{Introduction}

Despite their tremendous effectiveness in tasks such as object category detection, most deep neural networks do not understand the 3D nature of object categories. Reasoning about objects in 3D is necessary in many applications, for physical reasoning, or to understand the geometric relationships between different objects or scene elements.

The typical approach to learn 3D objects is to make use of large collections of high quality CAD models such as \cite{shapenet2015} or \cite{xiang2016objectnet3d}, which can be used to fully supervise models to recognize the objects' viewpoint and 3D shape. Alternatively, one can start from standard image datasets such as PASCAL VOC 
\cite{Everingham10}, augmented with other types of supervision, such as object segmentations and keypoint annotations~\cite{carreira16lifting}. Whether synthetically generated or manually collected, annotations have so far been required in order to overcome the significant challenges of learning 3D object categories, where both viewpoint and geometry are variable.

\begin{figure}[t]
\vspace{-1em}
\includegraphics[width=\linewidth,trim=0 1em 0 1em]{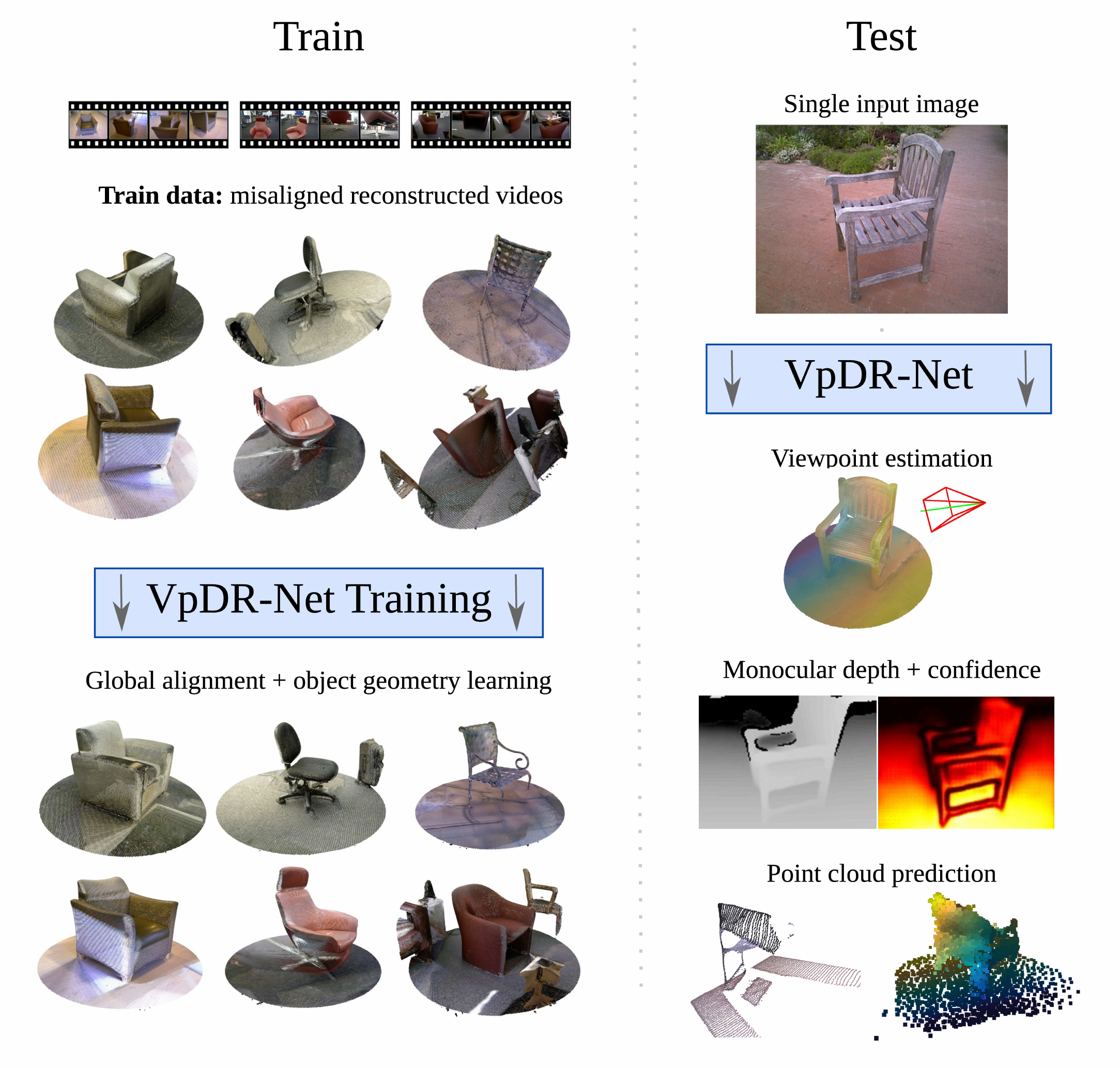}
\caption{We propose a convolutional neural network architecture to learn the 3D geometry of object categories from videos only, without manual supervision. Once learned, the network can predict i)~viewpoint, ii) depth, and iii) a point cloud, all from a single image of a new object instance.\label{fig:front}}
\end{figure}

In this paper, we develop an alternative approach that can learn 3D object categories in an \emph{unsupervised manner} (\cref{fig:front}), replacing synthetic or manual supervision with \emph{motion}. Humans learn about the visual word by experiencing it continuously, through a variable viewpoint, which provides very strong cues on its 3D structure. Our goal is to build on such cues in order to learn the 3D geometry of object categories, using videos rather than images of objects. We are motivated by the fact that videos are almost as cheap as images to capture, and do not require annotations.

\begin{figure*}
\includegraphics[width=\linewidth]{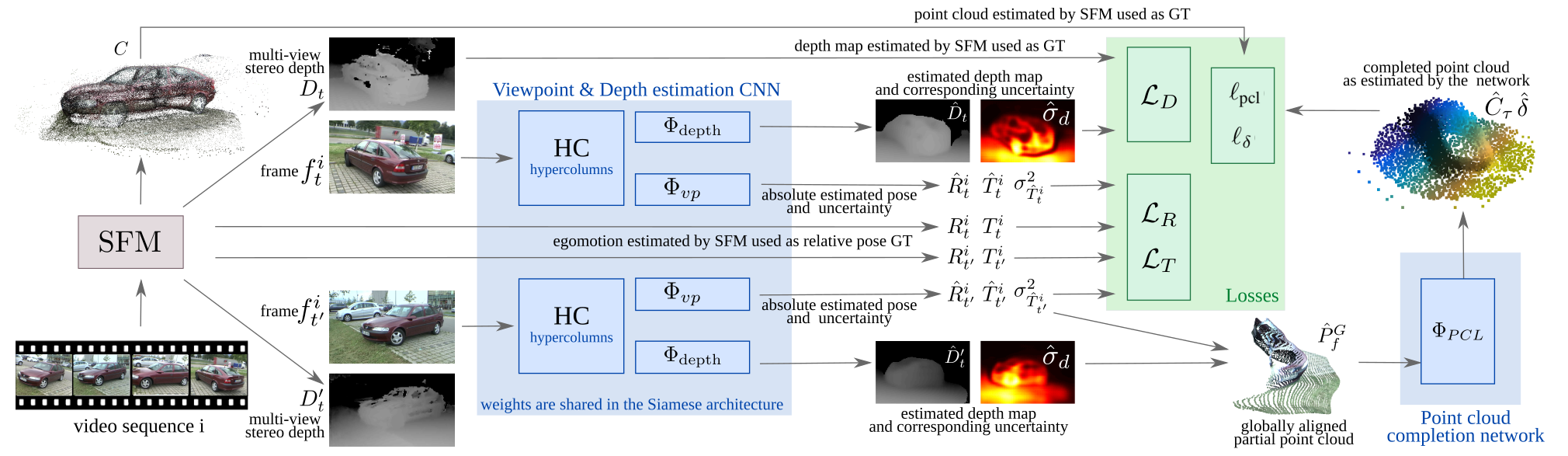}
\caption{\textbf{Overview of our architecture.} As a preprocessing, structure from motion (SFM) extracts egomotion and a depth map for every frame. For training, our architecture takes pairs of frames $f_t$, $f_{t'}$ and produces a viewpoint estimate, a depth estimate, and a 3D geometry estimate. At test time, viewpoint, depth, and 3D geometry are predicted from single images.\label{fig:overview}}
\end{figure*}

We build on mature structure-from-motion (SFM) technology to extract 3D information from individual video sequences. However, these cues are specific to each object instance as contained in different videos. The challenge is to integrate this information in a global 3D model of the object category, as well as to work with noisy and incomplete reconstructions from SFM. 

We propose a new deep architecture composed of three modules (\cref{fig:overview}). The first module estimates the \emph{absolute viewpoint} of objects in all video sequences (\cref{s:viewpoint}). This aligns different object instances to a common reference frame where geometric relationships can be modeled more easily. The second estimates the 3D shape of an object from a given viewpoint, producing a \emph{depth map} (\cref{s:depth}). The third \emph{completes the depth map to a full 3D reconstruction} in the globally-aligned reference frame (\cref{s:recon}). 
Combined and trained end-to-end without supervision, from videos alone, these components constitute \netname, 
a network for \textbf{v}iew\textbf{p}oint, \textbf{d}epth and \textbf{r}econstruction,
capable of extracting viewpoint and shape of a new object instance from a single image. 

One of our main contributions is thus to demonstrate the utility of using motion cues in learning 3D categories. We also introduce two significant technical innovations in the viewpoint and shape estimation modules as well as design guidelines and training strategies for 3D estimation tasks.

The first innovation (\cref{s:viewpoint}) is a new approach to align video sequences of different 3D objects based on a \emph{Siamese viewpoint factorization network}. While existing methods~\cite{sun09multi,sedaghat15unsupervised} align shapes by looking at 3D features, we propose to train \netname to directly estimate the absolute viewpoint of an object. We train our network to reconstruct \emph{relative camera motions}  and we show that this implicitly aligns different objects instances together. By avoiding explicit shape comparisons in 3D space, this method is simpler and more robust than alternatives. 

The second innovation (\cref{s:recon}) is a new network architecture that can generate a complete point cloud for the object from a partial reconstruction obtained from monocular depth estimation. This is based on a shape representation that predicts the support of a point probability distribution in 3D space, akin to a flexible voxelization, and a corresponding space occupancy map.

As a general design guideline, we demonstrate throughout the paper the utility of allowing deep networks to \emph{express uncertainty} in their estimate by predicting probability distributions over outputs (\cref{s:method}), yielding more robust training and useful cues (such as separating foreground and background in a depth map). We also demonstrate the significant power of \emph{geometry-aware data augmentation}, where a deep network is used to predict the geometry of an image and the latter is used to generate new realistic views to train other components of the system (\cref{s:augmentation}). Each component and design choice is thoroughly evaluated in~\cref{s:exp}, with significant improvements over the state-of-the-art.
\section{Related work}

\myparagraph{Viewpoint estimation} The vast majority of methods for learning the viewpoint of object categories use manual supervision~\cite{savarese073d,Ozuysal09pose,glasner11viewpoint,pepik14multi,xiang2014beyond,mottaghi2015coarse,tulsiani2015viewpoints} or synthetic~\cite{su2015render} data. In \cite{ummenhofer16demon}, a deep architecture predicts a relative camera pose and depth for a pair of images. Only a few works have used videos~\cite{sun09multi,sedaghat15unsupervised}. \cite{sedaghat15unsupervised} solves the shape alignment problem using a global search strategy based on the pairwise alignment of point clouds, a step we avoid by means of our Siamese viewpoint factorization network.

\myparagraph{3D shape prediction}
A traditional approach to 3D reconstruction is to use handcrafted 3D models~\cite{lawrence63machine, lowe87three}, and more recently 3D CAD models~\cite{shapenet2015,xiang2014beyond}. Often the  idea is to search for the 3D model in a CAD library that best fits the image~\cite{lim13parsing,aubry14seeing,gupta15aligning,bansal16marr}. Alternatively, CAD models can be used to train a network to directly predict the 3D shape of an object~\cite{girdhar16learning,wu16learning,tatarchenko16multi,choy163d}. Morphable models have sometimes been used ~\cite{zia13detailed,kar15category}, particularly for modeling faces~\cite{blanz03face,liu16joint}. All these methods require 3D models at train time.

\myparagraph{Data-driven approaches for geometry} Structure from motion (SFM) generally assumes fixed geometry between views and is difficult to apply directly to object categories due to intra-class variations. Starting from datasets of unordered images, methods such as~\cite{zhu10model} and \cite{prasad10finding} use SFM and manual annotations, such as keypoints in \cite{carreira16lifting,kar15category}, to estimate a rough 3D geometry of objects. Here, we leverage motion cues and do not need extra annotations.
\section{Method}\label{s:method}

We propose a single Convolutional Neural Network (CNN), \netname, that learns a \emph{3D object category} by observing it from a~\emph{variable viewpoint} in videos and no supervision (\cref{fig:overview}). Videos do not solve the problem of modeling intra-class shape variations, but they provide powerful yet noisy cues about the 3D shape of individual objects.

\netname takes as an input a set of $K$ video sequences $S^1, ..., S^K$ of an object category (such as cars or chairs), where a video $S^i = (f_1^i, ... , f_{N^i}^i)$ contains $N^i$ RGB or RGBD frames $f_t^i \in \mathbb{R}^{H \times W \times \mathcal{C} }$ (where $\mathcal{C}=3$ for RGB and $\mathcal{C}=4$ for RGBD data) 
and learns a model of the 3D category. This model has three components: i) a predictor $\PhiVP(f_i^t)$ of the \emph{absolute viewpoint} of the object (implicitly aligning the different object instances to a common reference frame; \cref{s:viewpoint}), ii) a \textit{monocular depth} predictor $\PhiDepth(f_i^t)$
(\cref{s:depth}) and iii) and a \textit{shape} predictor $\PhiPCL(f_i^t)$ that extends the depth map to a point cloud capturing the complete shape of the object (\cref{s:recon}). Learning starts by preprocessing videos to extract instance-specific egomotion and shape information (\cref{s:sfm}).

\subsection{Sequence-specific structure and pose}\label{s:sfm}

Video sequences are pre-processed to extract from each frame $f_t^i$ a tuple $(K_t^i,g_t^i,D_t^i)$ consisting of: (i) the camera calibration parameters $K_t^i$, (ii) its pose $g_t^i\in SE(3)$, and (iii) a depth map $D_t^i\in \mathbb{R}^{H \times W}$ associating a depth value to each pixel of $f_t^i$. The camera pose $g_t^i = (R_t^i,T_t^i)$ consists of a rotation matrix $R_t^i \in SO(3)$ and a translation vector $T_t^i \in \mathbb{R}^3$.\footnote{We use the convention that $g_t^i$ transforms world-relative coordinates $p_\text{world}$ to camera-relative coordinates $p_\text{camera} = g_t^i p_\text{world}$.}
We extract this information using off-the-shelf methods: the structure-from-motion (SFM) algorithm COLMAP for RGB sequences \cite{schoenberger2016sfm,schoenberger2016mvs}, and an open-source implementation~\cite{rusu2011pcl} of KinectFusion (KF) \cite{newcombe2011kinectfusion} for RGBD sequences. The information extracted from RGB or RGBD data is qualitatively similar, except that the scale of SFM reconstructions is arbitrary.

\subsection{Intra-sequence alignment}\label{s:viewpoint}

Methods such as SFM or KF can reliably estimate camera pose and depth information for single objects and individual video sequences, but are not applicable to~\emph{different instances and sequences}. In fact, their underlying assumption is that geometry is fixed, which is true for single (rigid) objects, but false when the geometry and appearance differ due to intra-class variations.

Learning 3D object categories requires to relate their variable 3D shapes by identifying and putting in correspondence analogous geometric features, such as the object front and rear. For rigid objects, such correspondences can be expressed by rigid transformations that \emph{align} occurrences of analogous geometric features.

The most common approach for aligning 3D shapes, also adopted by~\cite{sedaghat15unsupervised} for video sequences, is to extract and match 3D feature descriptors. Once objects in images or videos are aligned, the data can be used to supervise other tasks, such as learning a monocular predictor of the absolute viewpoint of an object~\cite{sedaghat15unsupervised}.

One of our main contributions, described below, is to reverse this process by learning a viewpoint predictor \emph{without} explicitly matching 3D shapes. Empirically (\cref{s:exp}), we show that, by skipping the intermediate 3D analysis, our method is often more effective and robust than alternatives.

\myparagraph{Siamese network for viewpoint factorization} Geometric analogies between 3D shapes can often be detected in image space directly, based on visual similarity. Thus, we propose to train a CNN $\PhiVP$ that maps a single frame $f_t^i$ to its \emph{absolute viewpoint} $\hat g_t^i = \PhiVP(f_t^i)$ in the globally-aligned reference frame. We wish to learn this CNN from the viewpoints estimated by the algorithms of~\cref{s:sfm} for each video sequence. However, these estimated viewpoints are \emph{not} absolute, but valid only within each sequence; formally, there are unknown sequence-specific motions $h^i = (R^i,T^i) \in SE(3)$ that map the sequence-specific camera poses $g^i_t$ to global poses $\hat g_t^i = g_t^i h^i$.\footnote{$h^i$ composes to the right: it transforms the world reference frame and then moves it to the camera reference frame.}

To address this issue, we propose to supervise the network using \emph{relative pose changes within each sequence}, which are invariant to the alignment transformation $h^i$. Formally, the transformation $h^i$ is eliminated by computing the relative pose change of the camera from frame $t$ to frame $t^\prime$:
\begin{equation}\label{e:fund}
   \hat g^i_{t^\prime} (\hat g^i_{t})^{-1}
   =
   g^i_{t^\prime} h^i (h^i)^{-1} (g^i_{t})^{-1}
   =
   g^i_{t^\prime} (g^i_{t})^{-1}.
\end{equation} 
Expanding the expression with $\hat g_t^i = (\hat R_t^i,\hat T_t^i)$, we find equations
expressing the relative rotation and translation
\begin{align}
\hat R_{t^\prime}^i (\hat R_t^i)^\top
&=
R_{t^\prime}^i (R_t^i)^\top,
\label{e:fund1}
\\
\hat T^i_{t^\prime} - \hat R_{t^\prime}^i (\hat R_t^i)^\top \hat T^i_{t}
&=
T^i_{t^\prime} - R_{t^\prime}^i (R_t^i)^\top T^i_{t}.
\label{e:fund2}
\end{align}
Eqs.~\eqref{e:fund1} and \eqref{e:fund2} are used to constrain the training of a \emph{Siamese architecture}, which, given two frames $t$ and $t'$, evaluates the CNN twice to obtain estimates $(\hat R^i_t, \hat T^i_t) = \PhiVP(f^i_t)$ and $(\hat R^i_{t^\prime}, \hat T^i_{t^\prime}) = \PhiVP(f^i_{t^\prime})$. The estimated poses are then compared to the ground truth ones, $(R^i_t,T^i_t)$ and $(R^i_{t^\prime},T^i_{t^\prime})$, in a relative manner by using losses that enforce the estimated poses to satisfy \cref{e:fund1,e:fund2}:
\begin{align}
\ell_R
(\hat R^i_t, \hat T^i_t, \hat R^i_{t^\prime}, \hat T^i_{t^\prime})
&\define
\|
\ln \hat R_{ t t^\prime}^i (R_{t t^\prime}^i)^\top
\|_F
\label{e:loss1}
\\
\ell_T (\hat R^i_t, \hat T^i_t, \hat R^i_{t^\prime}, \hat T^i_{t^\prime})
&\define
\|
\hat T_{t t^\prime}^i - T_{t t^\prime}^i
\|_2
\label{e:loss2}
\end{align}
where $\ln$ is the principal matrix logarithm and
\begin{align*}
R^i_{t^\prime t}
&\define
R_{t^\prime}^i (R_t^i)^\top,
&
\hat R^i_{t^\prime t}
&\define
\hat R_{t^\prime}^i (\hat R_t^i)^\top,
\\
T^i_{t^\prime t}
&\define
T^i_{t^\prime} - R^i_{t^\prime t} T^i_{t},
&
\hat T^i_{t^\prime t}
&\define
\hat T^i_{t^\prime} - \hat R^i_{t^\prime t} \hat T^i_{t}.
\end{align*}
While this CNN is only required to correctly predict relative viewpoint changes \emph{within each sequence}, since the \emph{same CNN} is used for all videos, the most plausible/regular solution for the network is to assign similar viewpoint predictions $(\hat R_t^i$, $\hat T_t^i)$ to images viewed from the same viewpoint, leading to a globally consistent alignment of the input sequences.  Furthermore, in a large family of 3D objects, different ones (e.g. SUVs and sedans) tend to be mediated by intermediate cases.
This is shown empirically in \cref{s:exp}.

\myparagraph{Scale ambiguity in SFM} \label{s:sfmambiguity} For methods such as SFM, there is an additional ambiguity: reconstructions are known only up to sequence-specific scaling factors $\lambda^i > 0$, so that the camera pose is parametrized as
$
 g^i_t(\lambda^i) = (R_t^i, \lambda^i T_t^i).
$
This ambiguity leaves~\cref{e:fund1} unchanged, but~\cref{e:fund2} becomes:
$$
\hat T^i_{t^\prime} - \hat R^i_{t^\prime t} \hat T^i_{t}
=
\lambda^i (T^i_{t^\prime} - R^i_{t^\prime t} T^i_{t})
\quad\Rightarrow\quad
\hat T^i_{t^\prime t} = \lambda^i T^i_{t^\prime t}
$$
During training, the ambiguity can be removed from loss~\eqref{e:loss2} by dividing vectors $T^i_{t^\prime t}$ and $\hat T^i_{t^\prime t}$ by their Euclidean norm. Note that for KF sequences $\lambda^i = 1$.
As the viewpoints are learned, an estimate of $\hat \lambda^i$ is computed using a moving average over training iterations for the other network modules to use (see supplementary material for details).

\myparagraph{Probabilistic predictions} Due to intrinsic ambiguities in the images or to errors in the SFM supervision (caused for example by reflective or textureless surfaces), $\PhiVP$ is occasionally unable to predict the ground truth viewpoint accurately. We found beneficial to allow the network to explicitly learn these cases and express uncertainty as an additional input-dependent prediction. For the translation component, we modify the network to predict the absolute pose $\hat T^i_t$ as well as its confidence score $\sigma_{\hat T^i_t}$ (predicted as the output of a soft ReLU units to ensure positivity). We then model the relative translation as a Gaussian distribution with standard deviation $\sigma_T = \sigma_{\hat T_{t^\prime}^i} + \sigma_{\hat T_{t}^i}$ and our model is now learned by minimizing the  negative log-likelihood $\mathcal{L}_T$ which replaces the loss $\ell_T$:
\begin{equation}\label{e:robust1}
\mathcal{L}_T =
-\ln \frac{1}{(2\pi\sigma^2_T)^\frac{3}{2}}
\exp\left(
-\frac{1}{2} \frac{\ell_T^2}{\sigma_T^2}
\right).
\end{equation}

The rotation component is more complex due to the non-Euclidean geometry of $SO(3)$, but it was found sufficient to assume that the error term~\eqref{e:loss1} has Laplace distribution and optimize
$
\mathcal{L}_R = - \ln \frac{1}{C_R} 
\exp\left(
-\frac{\sqrt{2} \ell_R}{\sigma_R}
\right),$
$\sigma_R =
\sigma_{\hat R_{t^\prime}^i}
+
\sigma_{\hat R_{t}^i},
$ where $C_R$ is a normalization term ensuring that the probability distribution integrates to one.
During training, by optimizing the losses $\mathcal{L}_R$ and $\mathcal{L}_T$ instead of $\ell_R$ and $\ell_T$, the network can discount gross errors by dividing the losses by a large predicted variance.

\myparagraph{Architecture} The architecture of $\PhiVP$ is a variant of ResNet-50~\cite{he16resnet} with some modifications to improve its performance as viewpoint predictor. The lower layers of $\PhiVP$ are used to extract a multiscale intermediate representation (denoted HC for hypercolumn \cite{hariharan2015hypercolumns} in \cref{fig:overview}). The upper layers consist of $2\times 2$ downsampling residual blocks that predict the viewpoint (see supp. material for details).

\begin{figure}
\centering
\newcommand{\augimagewidth}{0.157\linewidth}
\includegraphics[width=\augimagewidth]{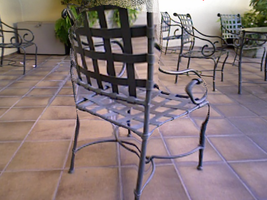}
\includegraphics[width=\augimagewidth]{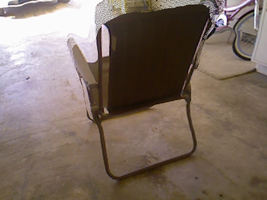}
\includegraphics[width=\augimagewidth]{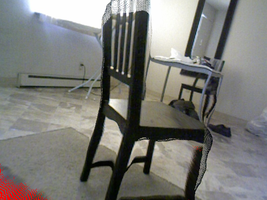}
\includegraphics[width=\augimagewidth]{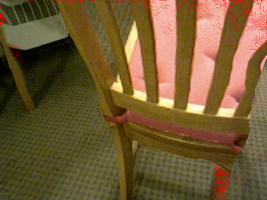}
\includegraphics[width=\augimagewidth]{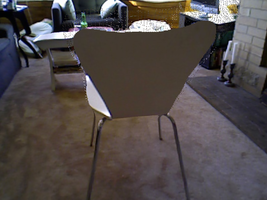}
\includegraphics[width=\augimagewidth]{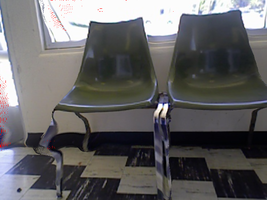}
\\
\includegraphics[width=\augimagewidth]{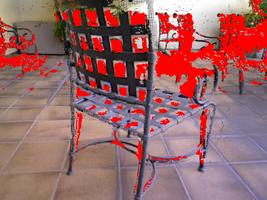}
\includegraphics[width=\augimagewidth]{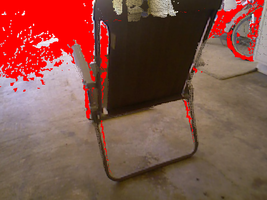}
\includegraphics[width=\augimagewidth]{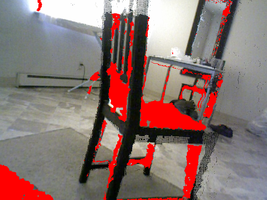}
\includegraphics[width=\augimagewidth]{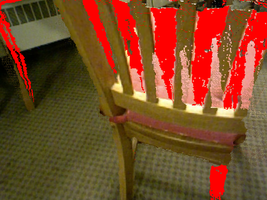}
\includegraphics[width=\augimagewidth]{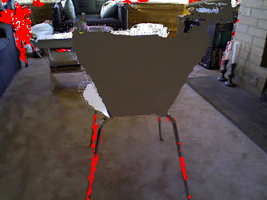}
\includegraphics[width=\augimagewidth]{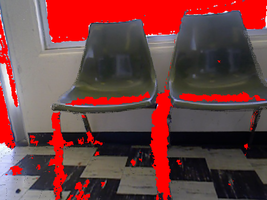}
\caption{\textbf{Data augmentation}. Training samples generated leveraging monocular depth estimation (ours, top) and using depth from KF (baseline, bottom).
  Missing pixels due to missing depth in red.}\label{f:aug}
\end{figure}

\subsection{Depth prediction}\label{s:depth}

The depth predictor module $\PhiDepth$ of \netname takes individual frames $f^i_t$ and outputs a corresponding depth map $\hat D_t = \PhiDepth(f^i_t)$, performing monocular depth estimation.

Estimating depth from a single image is inherently ambiguous and requires comparing the image to internal priors of the object shape. Similar to pose, we allow the network to explicitly \emph{learn and express uncertainty} about depth estimates by predicting a posterior distribution over possible pixel depths.
For robustness to outliers from COLMAP and KF, we assume a Laplace distribution
with negative log-likelihood loss
\begin{equation}
\mathcal{L}_D = \sum_{j=1}^{WH}
-\ln \frac{\sqrt{2}}{2\hat\sigma_{d_j}} ~
\exp\left(
-\frac{\sqrt{2} ~ |d_j - \hat {\lambda^i}^{-1} \hat d_j|}{\hat \sigma_{d_j} }
\right),
\end{equation}
where $d_j$ is the noisy ground truth depth output by SFM or KF for a given pixel $j$, and $\hat d_j$ and $\hat \sigma_{d_j}$ are respectively the corresponding predicted depth mean and standard deviation. The relative scale $\hat \lambda^i$ is 1 for KF and is estimated as explained in~\cref{s:sfmambiguity} for SFM.

\begin{figure*}
\newcommand{\vpimwidth}{1.5cm}
\includegraphics[width=\vpimwidth]{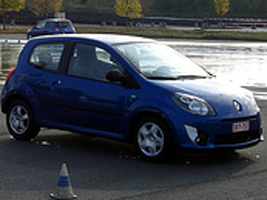}
\includegraphics[width=\vpimwidth]{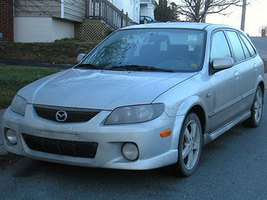}
\includegraphics[width=\vpimwidth]{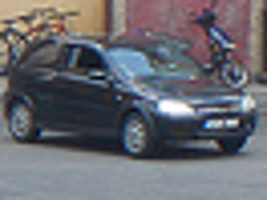}
\includegraphics[width=\vpimwidth]{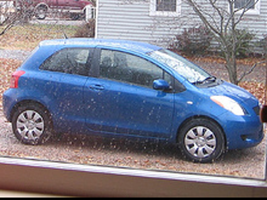}
\includegraphics[width=\vpimwidth]{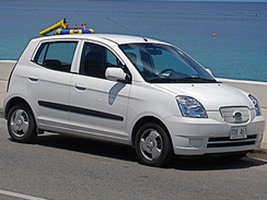}
\includegraphics[width=\vpimwidth]{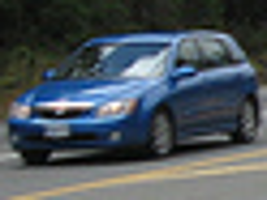}
\includegraphics[width=\vpimwidth]{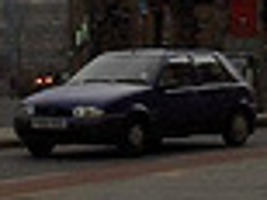}
\includegraphics[width=\vpimwidth]{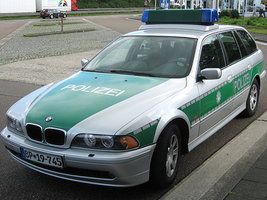}
\includegraphics[width=\vpimwidth]{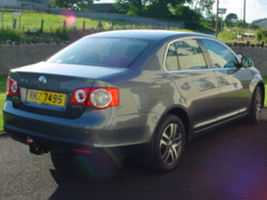}
\includegraphics[width=\vpimwidth]{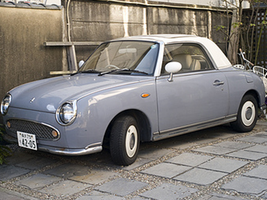}
\includegraphics[width=\vpimwidth]{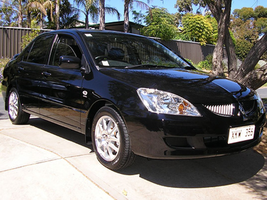}
\\
\includegraphics[width=\vpimwidth,trim=0 13pt 0 13pt,clip]{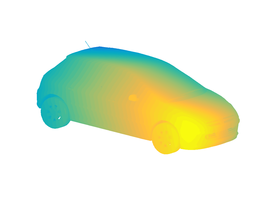}
\includegraphics[width=\vpimwidth,trim=0 13pt 0 13pt,clip]{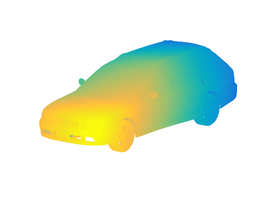}
\includegraphics[width=\vpimwidth,trim=0 13pt 0 13pt,clip]{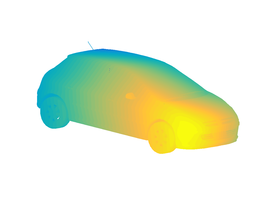}
\includegraphics[width=\vpimwidth,trim=0 13pt 0 13pt,clip]{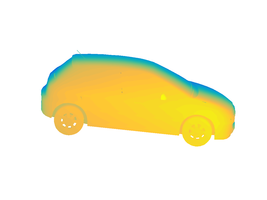}
\includegraphics[width=\vpimwidth,trim=0 13pt 0 13pt,clip]{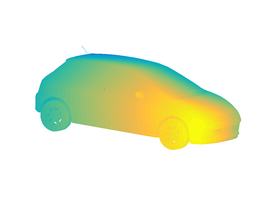}
\includegraphics[width=\vpimwidth,trim=0 13pt 0 13pt,clip]{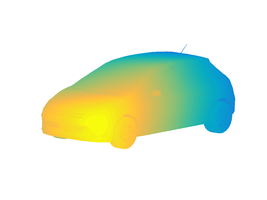}
\includegraphics[width=\vpimwidth,trim=0 13pt 0 13pt,clip]{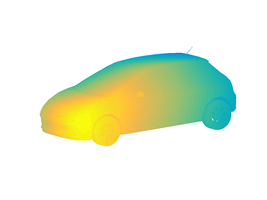}
\includegraphics[width=\vpimwidth,trim=0 13pt 0 13pt,clip]{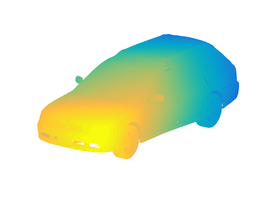}
\includegraphics[width=\vpimwidth,trim=0 13pt 0 13pt,clip]{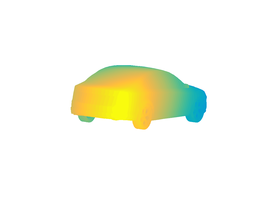}
\includegraphics[width=\vpimwidth,trim=0 13pt 0 13pt,clip]{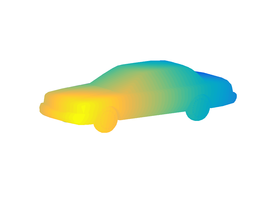}
\includegraphics[width=\vpimwidth,trim=0 13pt 0 13pt,clip]{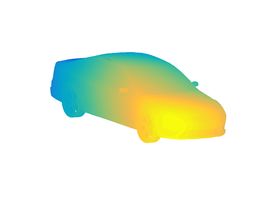}
\\
\includegraphics[width=\vpimwidth]{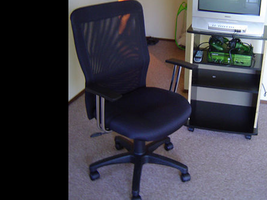}
\includegraphics[width=\vpimwidth]{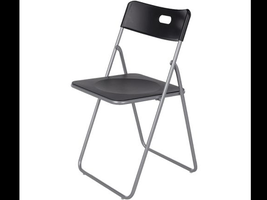}
\includegraphics[width=\vpimwidth]{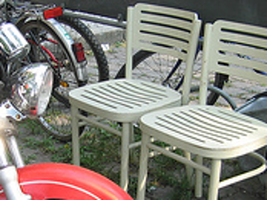}
\includegraphics[width=\vpimwidth]{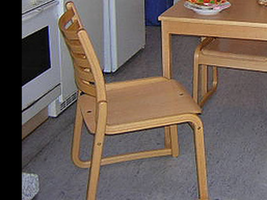}
\includegraphics[width=\vpimwidth]{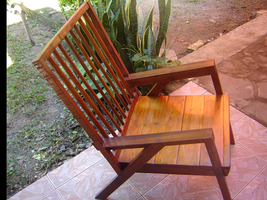}
\includegraphics[width=\vpimwidth]{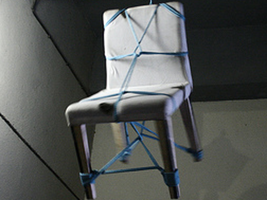}
\includegraphics[width=\vpimwidth]{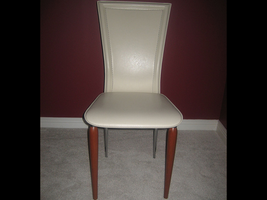}
\includegraphics[width=\vpimwidth]{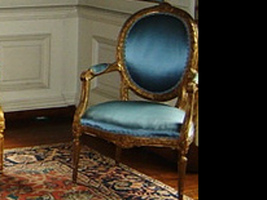}
\includegraphics[width=\vpimwidth]{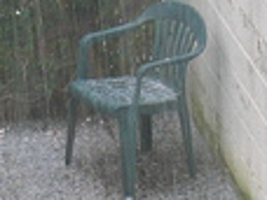}
\includegraphics[width=\vpimwidth]{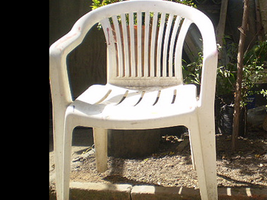}
\includegraphics[width=\vpimwidth]{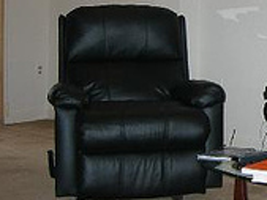}
\\
\includegraphics[width=\vpimwidth,trim=0 6pt 0 6pt,clip]{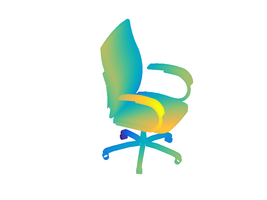}
\includegraphics[width=\vpimwidth,trim=0 6pt 0 6pt,clip]{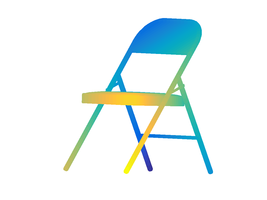}
\includegraphics[width=\vpimwidth,trim=0 6pt 0 6pt,clip]{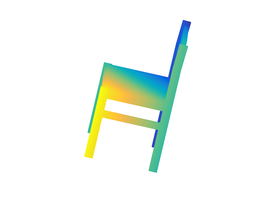}
\includegraphics[width=\vpimwidth,trim=0 6pt 0 6pt,clip]{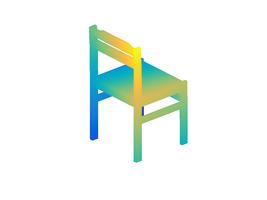}
\includegraphics[width=\vpimwidth,trim=0 6pt 0 6pt,clip]{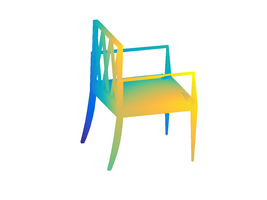}
\includegraphics[width=\vpimwidth,trim=0 6pt 0 6pt,clip]{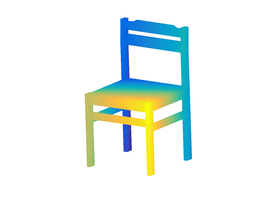}
\includegraphics[width=\vpimwidth,trim=0 6pt 0 6pt,clip]{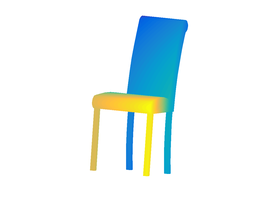}
\includegraphics[width=\vpimwidth,trim=0 6pt 0 6pt,clip]{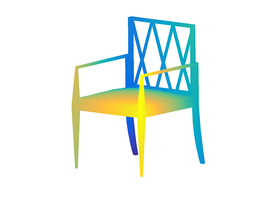}
\includegraphics[width=\vpimwidth,trim=0 6pt 0 6pt,clip]{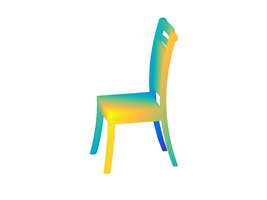}
\includegraphics[width=\vpimwidth,trim=0 6pt 0 6pt,clip]{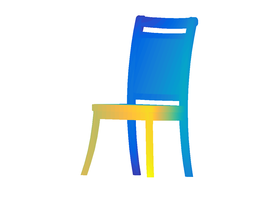}
\includegraphics[width=\vpimwidth,trim=0 6pt 0 6pt,clip]{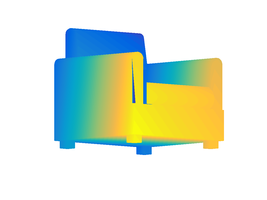} 
\caption{\textbf{Viewpoint prediction.} Most confident viewpoint predictions (sorted by predicted confidence from left to right) where the viewpoint predicted by \netname is used to align the Pascal3D ground truth CAD models with each image. \label{fig:vp_qual}}
\vspace{-0.2cm}
\end{figure*}

\subsection{Point-cloud completion}\label{s:recon}

Given any image $f$ of an object instance, its \textit{aligned 3D shape} can be reconstructed by estimating and aligning its depth map using the output of the viewpoint and depth predictors of~\cref{s:depth,s:viewpoint}. However, since a depth map cannot represent the occluded portions of the object, such a reconstruction can only be partial. In this section, we describe the third and last component of~\netname, whose goal is to generate a full reconstruction of the object, beyond what is visible in the given view.

\myparagraph{Partial point cloud} The first step is to convert the predicted depth map $\hat D_f = \PhiDepth(f)$ into a partial point cloud
$
\hat P_f
\define
\{ \hat p_j :  j=1,\dots,HW \},
$
$
\hat p_j
\define
K^{-1}
\begin{bmatrix} u_j & v_j & \hat d_i \end{bmatrix}^\top,
$
where $(u_j,v_j)$ are the coordinates of a pixel $j$ in the depth map $\hat D_f$ and $K$ is the camera calibration matrix. Empirically, we have found that the reconstruction problem is much easier if the data is aligned in the global reference frame established by \netname. Thus, we transform $\hat P_f$ into a globally-aligned point cloud as $\hat P^G_f = \hat g^{-1} \hat P_f$, where $\hat g=\PhiVP(f)$ is the camera pose estimated by the viewpoint-prediction network.

\myparagraph{Point cloud completion network} Next, our goal is to learn the point cloud completion part of our network $\PhiPCL$ that takes the aligned but incomplete point could $\hat P^G_f$ and produces a complete object reconstruction $\hat C$. We do so by predicting a 3D occupancy probability field. However, rather than using a volumetric method that may require a discrete and fixed voxelization of space, we propose a simple and efficient alternative. First, the network $\PhiPCL$ predicts a  set of $M$ 3D points $\hat S = (\hat s_1,\dots, \hat s_M) \in \mathbb{R}^{3\times M}$ that, during training, closely fit the ground truth 3D point cloud $C$. This step minimizes the fitting error:
\begin{equation}
 \lossPCL(\hat S) = \frac{1}{|C|}\sum_{c \in C}
 \min_{m=1,\dots,M} \left\| c -  \hat s_m \right\|_2.
\end{equation}
The 3D point cloud $\hat S$ provides a good coverage of the ground truth object shape. However, this point cloud is conservative and distributed \emph{in the vicinity} of the ground truth object. Thus, while this is not a precise representation of the object shape, it works well as a support of a probability distribution of space occupancy. In order to estimate the occupancy probability values, the network $\PhiPCL(\hat P^G_f)$ predicts additional scalar outputs
$$
\delta_m = | \{ c \in C : \forall m^\prime: \| \hat s_m - c \|_2 \leq \| \hat s_{m^\prime} - c \|_2 \} | / |C|
$$
proportional to the number of  ground truth surface points $c \in C$ for which the support point $\hat s_m$ is the nearest neighbor. The network is trained to compute a prediction $\hat\delta_m$ of the occupancy masses $\delta_m$ by minimizing the squared error loss
$
\ell_\delta(\hat \delta, \delta) = \sum_{m=1}^M (\hat \delta_m - \delta_m)^2.
$

Given the network prediction $(\hat S, \hat \delta) = \PhiPCL(\hat P^G_f)$, the completed point cloud is then defined as the subset of points $\hat C$ that have sufficiently high occupancy, defined as:
$
\hat C_\tau = \{ \hat s_m \in \hat S : \delta_m \geq \tau \}
$
where $\tau$ is a confidence parameter. The set $\hat C_\tau$ can be further refined by using e.g.\ a 3D Laplacian filter to smooth out noise.

\myparagraph{Architecture} The point cloud completion network $\PhiPCL$ is modeled after PointNet \cite{qi16pointnet}, originally proposed to semantically \emph{segment} a point clouds. Here we adapt it to perform a completely different task, namely 3D shape reconstruction. This is made possible by our model where shape is represented as a cloud of 3D support points $\hat S$ and their occupancy masses $\hat \delta$.
Differently from $\PhiVP$ and $\PhiDepth$, $\PhiPCL$ is \emph{not} convolutional but uses a sequence of fully connected layers to process the 3D points in $\hat P_f^G$, after appending an appearance descriptor to each of them. A key step is to add an intermediate orderless pooling operator to remove the dependency on the order and number of input points (see the supplementary material for details). The architecture is configured to predict $M=10^4$ points $\hat S$.
\begin{table*}[t!]
\newcommand{\methodbox}[1]{#1}
\centering \scriptsize
\begin{tabular}{llllcccccc}
  \toprule
\textbf{object class}             & \textbf{test set} & \textbf{level of supervision} & \textbf{method} & $\downarrow$ $e_R$ & $\downarrow$ $e_C$ & $\downarrow$ $e_R^{rel}$ & $\downarrow$ $e_T^{rel}$ & $\uparrow AP_{e_R}$ & $\uparrow AP_{e_C}$ \\  \midrule
\multirow{3}{*}{car}   &  \multirow{3}{*}{Pascal3D}  & unsupervised & \methodbox{VPNet} + aligned FrC \cite{sedaghat15unsupervised}  & 49.62 & 32.29 & 85.45 & 0.84 & 0.15 & 0.01 \\ 
                       &                           & unsupervised & \textbf{\methodbox{\netname} + FrC  (ours)}                              & \textbf{29.57} & \textbf{7.29}  & \textbf{62.30} & \textbf{0.65} & \textbf{0.41} & \textbf{0.91} \\ \cmidrule{3-10 }
                       &  & fully supervised & \methodbox{VPNet} + Pascal3D                                   & 12.49 & 1.27  & 20.34 & 0.24 & 0.77 & 0.97 \\  
\midrule
\multirow{6}{*}{chair}  & \multirow{3}{*}{Pascal3D}         & unsupervised & \methodbox{VPNet} + aligned LDOS \cite{sedaghat15unsupervised} & 64.68 & 42.46 & 89.01 & 0.95 & 0.06 & 0.00 \\ 
&                           & unsupervised & \textbf{\methodbox{\netname} + LDOS (ours) }                             & \textbf{42.34} & \textbf{16.72} & \textbf{71.35} & \textbf{0.93} & \textbf{0.23} & \textbf{0.22} \\\cmidrule{3-10 }
						&	& fully supervised & \methodbox{VPNet} + Pascal3D                                  & 34.37 & 6.14  & 67.41 & 0.74 & 0.26 & 0.66\\
\cmidrule{2-10}
& \multirow{3}{*}{LDOS}                            & unsupervised & \methodbox{VPNet} + aligned LDOS \cite{sedaghat15unsupervised} & \textbf{30.56} & 0.61  & 71.40 & 0.77 & 0.30 & 0.18 \\ 
                       &                           & unsupervised & \textbf{\methodbox{\netname} + LDOS (ours) }                             & 33.92 & \textbf{0.54}  & \textbf{60.90} & \textbf{0.70} & \textbf{0.40} & \textbf{0.22} \\  \cmidrule{3-10 } 
                       & & fully supervised & \methodbox{VPNet} + Pascal3D                              & 61.45 & 2.55  & 82.97 & 0.96 & 0.15 & 0.00\\
\bottomrule
\end{tabular}
\caption{\textbf{Viewpoint prediction.} Angular error $e_r$ and camera-center distance $e_c$ for absolute pose evaluation, and relative camera rotation error $e_R^{rel}$ and translation error $e_T^{rel}$ for relative pose evaluation. $AP_{e_R}$ and $AP_{e_C}$ evaluate absolute angular error and camera-center distance of the pose predictions taking into account the associated estimate confidence values. \netname trained on video sequences, is compared to VPNet trained on aligned video sequences and a fully-supervised VPNet. $\uparrow$ (resp. $\downarrow$) means larger (resp. lower) is better.}
\label{tab:posest}
\vspace{-1em}
\end{table*}

\myparagraph{Leave out} During training the incomplete point cloud $\hat P_f^G$ is downsampled by randomly selecting between $M = 10^3$ and $10^4$ points based on their depth prediction confidence as estimated by $\PhiDepth$. Similar to dropout, dropping points allows the network to overfit less, to become less sensitive to the size of the input point cloud, and to implicitly discard background points (as these are assigned low confidence by depth prediction). For the latter reason, leave out is maintained at test time too with $M=10^4$.
\section{Geometry-aware data augmentation}\label{s:augmentation}

As viewpoint prediction with deep networks benefits significantly from large training sets~\cite{su2015render}, we increase the effective size of the training videos by \emph{data augmentation}. This is trivial for tasks such as classification, where one can translate or scale an image without changing its identity. The same is true for viewpoint recognition if the task is to only estimate the viewpoint orientation as in~\cite{su2015render,tulsiani2015viewpoints}, as images can be scaled and translated without changing the equivalent viewpoint orientation. However, this assumption is not satisfied if, as in our case, the goal is to estimate all 6 DoF of the camera pose.

Inspired by the approach of~\cite{gupta2016synthetic}, we propose to solve this problem by using the estimated scene geometry to \emph{generate new realistic viewpoints} (\cref{f:aug}). Given a sample $(f^i_t, g^i_t, D^i_t)$, we apply a random perturbation to the viewpoint (with a forward bias to avoid unoccluding too many pixels) and use depth-image-based rendering (DIBR)~\cite{morvan2009acquisition} to generate a new sample $(f^i_*, g^i_*, D^i_*)$, warping both the image and the depth map.

Sometimes the depth map $D^i_t$ from KF contains too many holes to yield satisfactory DIBR results (\cref{f:aug}, bottom); we found preferable to use the depth $\hat D^i_t = \PhiDepth(f_t)$ estimated by the network which is less accurate but more robust, containing almost no missing pixels (\cref{f:aug}, top).

\begin{table}[t!]
\centering
\newcommand{\testsetbox}[1]{#1}
\centering \scriptsize
\setlength\tabcolsep{2pt}
\begin{tabular}{lp{0.95cm}p{0.95cm}p{0.95cm}p{0.95cm}p{0.95cm}p{0.95cm}}
 \toprule
                & $\downarrow$ $e_R$ & $\downarrow$ $e_C$ & $\downarrow$ $e_R^{rel}$ & $\downarrow$ $e_T^{rel}$ & $\uparrow$ $AP_{e_R}$ & $\uparrow$ $AP_{e_C}$ \\
 \midrule
               & \multicolumn{6}{c}{Test set: \testsetbox{\textbf{LDOS}} }  \\
\midrule
\textbf{\netname (ours)}  & \tb{33.92} & \tb{0.54}  & \tb{60.90} & \tb{0.70}  & \tb{0.40} & \tb{0.22} \\
\netname-NoProb  & 45.33 & 0.67  & 69.33 & 0.85  & 0.12 & 0.07 \\
\netname-NoDepth & 68.19 & 0.85  & 82.99 & 1.01  & 0.01 & 0.01 \\
\netname-NoAug   & 35.16 & 0.59  & 63.54 & 0.73  & 0.38 & 0.19 \\
\midrule
               & \multicolumn{6}{c}{Test set: \testsetbox{\textbf{Pascal3D}} } \\ 
\midrule
\textbf{\netname (ours)}  & \tb{42.34} & \tb{16.72} & \tb{71.35} & 0.93  & \tb{0.23} & \tb{0.22} \\
\netname-NoProb  & 57.23 & 17.06 & 77.72 & 1.05  & 0.08 & 0.14 \\
\netname-NoDepth & 60.31 & 17.89 & 85.17 & 1.15  & 0.07 & 0.21 \\
\netname-NoAug   & 43.52 & 18.80 & 72.93 & \tb{0.92}  & 0.10 & 0.17 \\
\bottomrule
\end{tabular}
\caption{\textbf{Viewpoint prediction.} Different flavors of \netname with removed components to evaluate their respective impact. 
  \label{tab:ablation}}
\vspace{-1em}
\end{table}


\section{Experiments}\label{s:exp}

We assess viewpoint estimation in~\cref{s:exp-pose}, depth prediction in~\cref{s:exp-depth}, and point cloud prediction in~\cref{s:exp-pcc}.

\myparagraph{Datasets} Throughout the experimental section, we consider three datasets for training and benchmarking our network: (1) \textbf{FreiburgCars (FrC)}~\cite{sedaghat15unsupervised}  which consists of RGB video sequences with the camera circling around various types of cars; (2) the \textbf{Large Dataset of Object Scans (LDOS)}~\cite{choi2016large} containing RGBD sequences of man-made objects; and (3) \textbf{Pascal3D}~\cite{xiang2014beyond}, a standard benchmark for pose estimation~\cite{tulsiani2015viewpoints,su2015render}.

For viewpoint estimation, Pascal3D already contains viewpoint annotations. For LDOS, experiments focus on the \textit{chair} class. In order to generate ground truth pose annotations for evaluation, we manually aligned 3D reconstructions of 10 randomly-selected chair videos and used 50 randomly-selected frames for each video as a test set. 

For depth estimation, we evaluate on LDOS as it provides high quality depth maps one can use as ground truth.

For point cloud reconstruction, we use FrC and LDOS. Ground truth point clouds for evaluation are obtained by merging the SFM or RGBD depth maps from all frames of a given test video sequence, sampling $3\cdot10^4$ points and post-processing those using a 3D Laplacian filter. For FrC, five videos were randomly selected and removed from the train set, picking 60 random frames per video for evaluation. For LDOS the pose estimation test frames are used.

\myparagraph{Learning details} \netname is trained with stochastic gradient descent with a momentum of 0.0005 and an initial learning rate of $10^{-2}$. The weights of the losses were empirically set to achieve convergence on the training set. 
Better convergence was observed by training \netname in two stages. First, $\PhiDepth$ and $\PhiVP$ were optimized jointly, lowering the learning rate tenfold when no further improvement in the training losses was observed. Then, $\PhiPCL$ is optimized after initializing the bias of its last layer, which corresponds to an average point cloud of the object category, by randomly sampling points from the ground truth models.

\subsection{Pose estimation}\label{s:exp-pose}

\myparagraph{Pascal3D} First, we evaluate the \netname viewpoint predictor on the Pascal3D benchmark \cite{xiang2014beyond}. Unlike previous works \cite{su2015render,tulsiani2015viewpoints} that focus on estimating the object/camera viewpoint represented by a 3 DoF rotation matrix, we evaluate the full 6 DoF camera pose represented by the rotation matrix $R$ together with the translation vector $T$.

In Pascal3D, the camera poses are expressed relatively to the whole scenes instead of the objects themselves, so we adjust the dataset annotations. 
We crop every object using bounding box annotations after reshaping the box to a fixed aspect ratio, and resize the crop to $240\times 320$ pixels. The camera pose is adjusted to the cropped object using the P3P algorithm to minimize the reprojection error between the camera-projected vertices of the ground truth CAD model and the original projection after cropping and resizing.

\myparagraph{Absolute pose evaluation} We first evaluate absolute camera pose estimation using two standard measures: the angular error
$
e_R = 2^{-\frac{1}{2}}
\|\ln R^\ast \hat R^\top \|_F
$ 
between the ground truth camera pose $R^\ast$ and the prediction $\hat R$ \cite{tulsiani2015viewpoints,su2015render}, as well as the camera-center distance
$
e_C = \|\hat C - C^\ast\|_2
$
between the predicted camera center $\hat C$ and the ground truth $C^\ast$. Following the common practice \cite{tulsiani2015viewpoints,su2015render} we report median $e_R$ and $e_C$ over all pose predictions on each test set.

Note that, while object viewpoints in Pascal3D and our method are internally consistent for a whole category, they may still differ between them by an arbitrary global 3D similarity transformation. Thus, as detailed in the supplementary material, the two sets of annotations are aligned by a single global similarity $\mathcal{T}_G$ before assessment.

\myparagraph{Relative pose evaluation} To assess methods with measures independent of $\mathcal{T}_G$ we also evaluate: (1) the relative rotation error between pairs of ground truth relative camera motions $R_{t t^\prime}^\ast$ and the corresponding predicted relative motions $\hat R_{t t^\prime}$ given by
$
e_R^\text{rel} = 
2^{-\frac{1}{2}}
\|\ln
R_{t t^\prime}^\ast
\hat R_{t t^\prime}^\top
\|_F
$
and (2) the normalized relative translation error 
$
e_T^\text{rel}
= \|
\hat T_{t t^\prime} - T_{t t^\prime}^\ast
\|_2
$%
, where both $\hat T_{t t^\prime}$ and $T_{ t t^\prime}^\ast$ are $\ell_2$-normalized so the measure is invariant to the scaling component of $\mathcal{T}_G$. We report the median errors over all possible image pairs in each test set.

\begin{figure}[t] 
	\centering
   \pbox{\textwidth}{
   \includegraphics [width=0.47\linewidth] {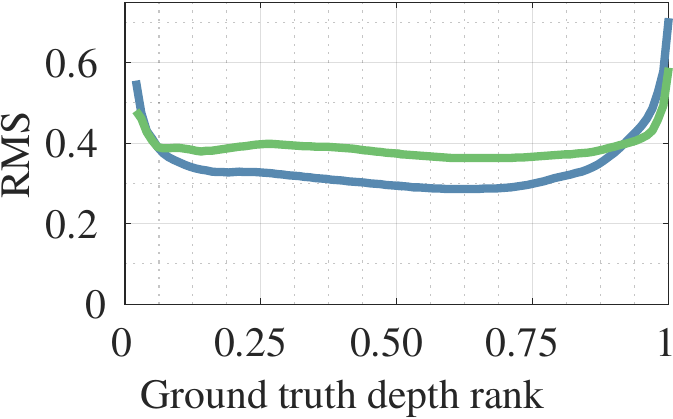}
   \includegraphics [width=0.48\linewidth] {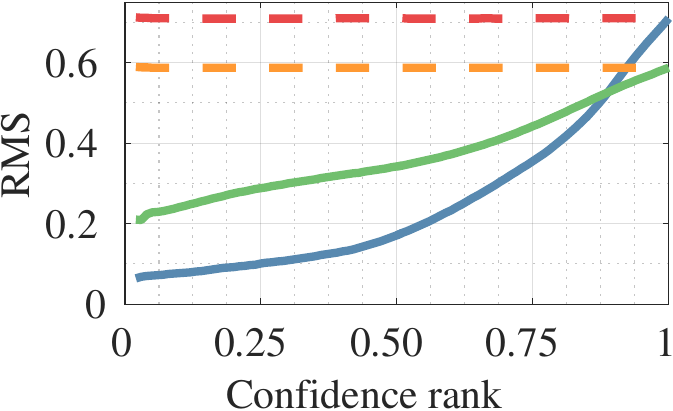}\\
   \includegraphics [width=1\linewidth] {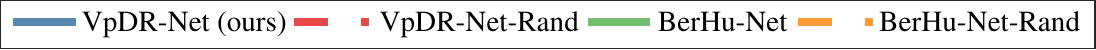}
   }
  \caption{\textbf{Monocular depth prediction.} Cumulative RMS depth reconstruction error for the LDOS data, when pixels are ranked by ground truth depth (left) and by confidence (right).}
  \label{fig:depthest}
\end{figure}

\begin{figure}[t]
\newcommand{\depthimheight}{0.84cm}
\centering
\includegraphics[height=\depthimheight]{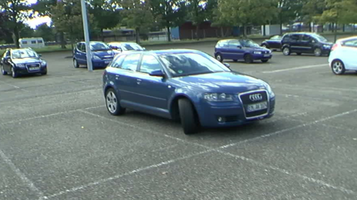}
\includegraphics[height=\depthimheight]{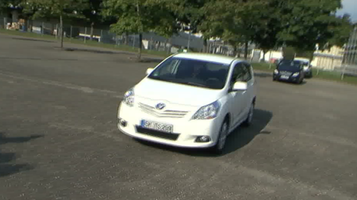}
\includegraphics[height=\depthimheight]{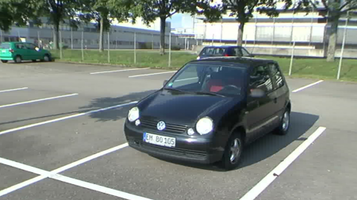}
\includegraphics[height=\depthimheight]{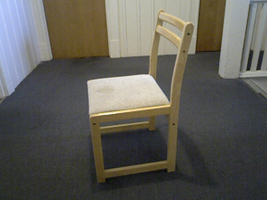}
\includegraphics[height=\depthimheight]{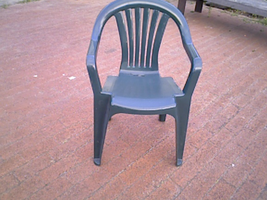}
\includegraphics[height=\depthimheight]{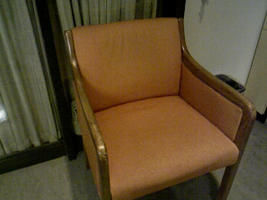}\\
\includegraphics[height=\depthimheight]{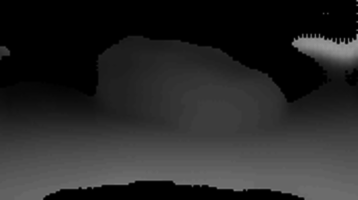}
\includegraphics[height=\depthimheight]{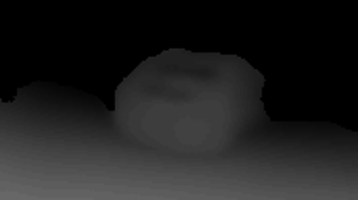}
\includegraphics[height=\depthimheight]{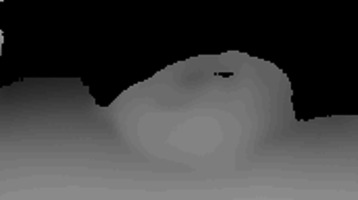}
\includegraphics[height=\depthimheight]{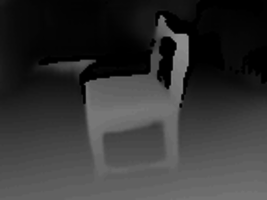}
\includegraphics[height=\depthimheight]{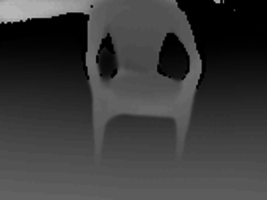}
\includegraphics[height=\depthimheight]{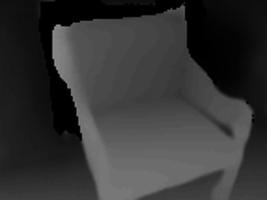}\\
\includegraphics[height=\depthimheight]{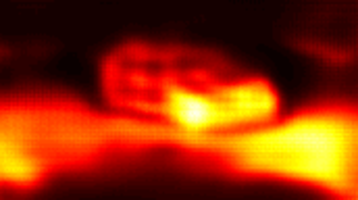}
\includegraphics[height=\depthimheight]{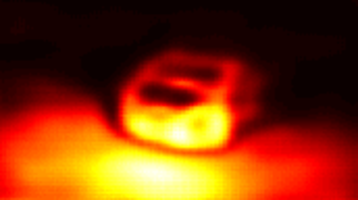}
\includegraphics[height=\depthimheight]{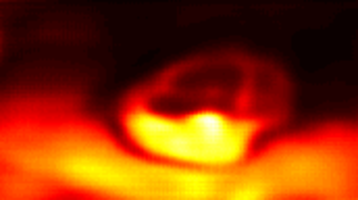}
\includegraphics[height=\depthimheight]{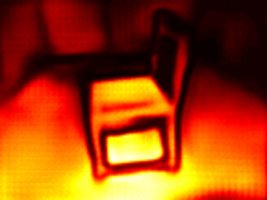}
\includegraphics[height=\depthimheight]{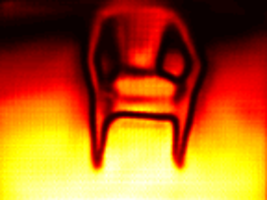}
\includegraphics[height=\depthimheight]{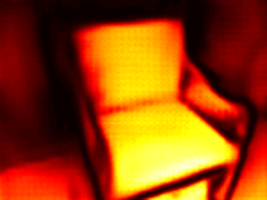}\\
\caption{\textbf{Monocular depth prediction.} Top: input image; middle: predicted depth; bottom: predicted depth confidence.
Depth maps are filtered by removing low confidence pixels.
\label{fig:depth_qual}}
\vspace{-1em}
\end{figure}

\myparagraph{Pose prediction confidence evaluation} A feature of our model is to produce confidence scores with its viewpoint estimates. We evaluate the reliability of these scores by correlating them with viewpoint prediction accuracy. In order to do so, predictions are divided into ``accurate'' and ``inaccurate'' by comparing their errors $e_R$ and $e_C$ to thresholds (set to $e_R=\frac{\pi}{6}$ following~\cite{su2015render,tulsiani2015viewpoints} and $e_C=15$ and $0.5$ for Pascal3D or LDOS respectively). Predictions are then ranked by decreasing confidence scores and the average precisions $AP_{e_R}$ and $AP_{e_C}$ of the two ranked lists are computed. 

\begin{figure*}[t]
\centering
\includegraphics[width=\linewidth]{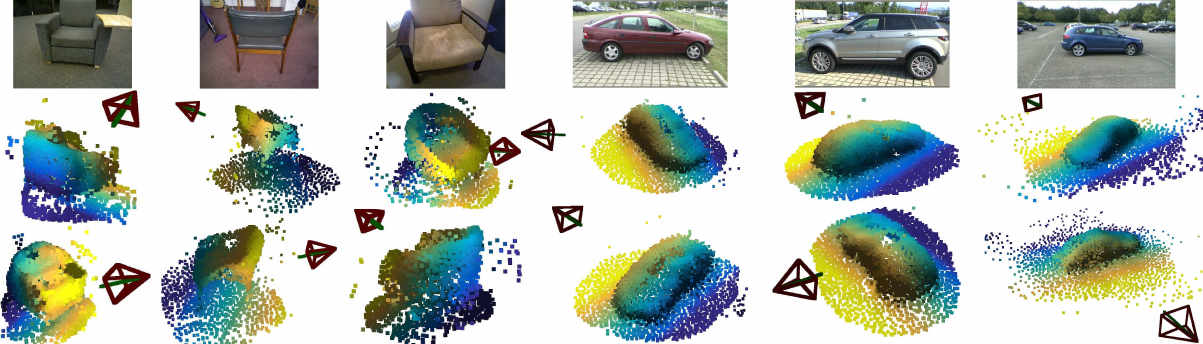}
\caption{\textbf{Point cloud prediction.} From a single input image of an unseen object instance (top row), \netname predicts the 3D geometry of that instance in the form of a 3D point cloud (seen from two different angles, middle and bottom rows).\label{fig:pcl_qual}}
\vspace{-1em}
\end{figure*}

\myparagraph{Baselines} We compare our viewpoint predictor to a strong baseline, called~\textbf{VPNet}, trained using absolute viewpoint labels. VPNet is a ResNet50 architecture \cite{he16resnet} with the final softmax classifier replaced by a viewpoint estimation layer that predicts the 6 DoF pose $\hat g_t^i$. Following \cite{tulsiani2015viewpoints}, rotation matrices are decomposed in Euler angles, each discretized in 24 equal bins.
This network is trained to predict a softmax distribution over the angular bins and to regress a 3D vector corresponding to the camera translation $T$. The average softmax value across the three max-scoring Euler angles is used as a prediction confidence score.

We test both an unsupervised and a fully-supervised variant of VPNet. VPNet-unsupervised is comparable to our setting and is trained on the output of the global camera poses estimated from the videos by the state-of-the-art sequence-alignment method of \cite{sedaghat15unsupervised}. In the fully-supervised setting, VPNet is trained instead by using ground-truth global camera poses provided by the Pascal3D training set.

\myparagraph{Results} \Cref{tab:posest} compares \netname to the VPNet baselines. First, we observe that our baseline VPNet-unsupervised is very strong, as we report $e_R=49.6$ error for the full rotation matrix, while the original method of \cite{sedaghat15unsupervised} reports an error of 61.5 just for the azimuth component. Nevertheless, \netname outperforms VPNet in all performance metrics except for a single case ($e_R$ for LDOS chairs). Furthermore, the advantage is generally substantial,  and the unsupervised \netname reduces the gap with fully-supervised VPNet by 20 \% or better in the vast majority of the cases. This shows the advantage of the proposed viewpoint factorization method compared  to aligning 3D shapes as in~\cite{sedaghat15unsupervised}. Second, we observe that the confidence scores estimated by \netname are significantly more correlated with the accuracy of the predictions than the softmax scores in VPNet, providing a reliable self-assessment mechanism. The most confident viewpoint predictions of \netname are shown in \cref{fig:vp_qual}. 

\myparagraph{Ablation study} We evaluate the importance of the different components of \netname by turning them off and measuring performance on the \textit{chair} class. In \cref{tab:ablation}, \textbf{\netname{}-NoProb} replaces the robust probabilistic losses $\mathcal{L}_R$ and $\mathcal{L}_T$ with their non-probabilistic counterparts $\ell_R$ and $\ell_T$, and confidence predictions are replaced with random scores for AP evaluation. \textbf{\netname{}-NoDepth} removes the depth prediction and point cloud prediction branches during training, retaining only the $\PhiVP$ subnetwork. \textbf{\netname{}-NoAug} does not use the data augmentation mechanism of \cref{s:augmentation}.

We observe a significant performance drop when each of the components is removed. This confirms the importance of all contributions in the network design. 
Interestingly, we observe that the depth prediction branch $\PhiDepth$ is  crucial for pose estimation (\eg~-34.27 $e_R$ on LDOS). 

\begin{table}
  \centering 
  \footnotesize
  \setlength\tabcolsep{2pt}
\begin{tabular}{lrrrr}
\toprule
Test set                          & \multicolumn{2}{c}{\textbf{LDOS}} & \multicolumn{2}{c}{\textbf{FrC}} \\
\midrule
Metric                            &  $\uparrow$ mVIoU    & $\downarrow$ m$D_{pcl}$  & $\uparrow$ mVIoU     & $\downarrow$ m$D_{pcl}$   \\
\midrule
Aubry \cite{aubry14seeing}       & 0.06      & 1.30        & 0.21      & 0.41                 \\
\textbf{\netname (ours) }         & \textbf{0.13}      & 0.20        & 0.24      & 0.28                 \\
\textbf{\netname-Fuse (ours)}    & \textbf{0.13}      & \textbf{0.19}        & \textbf{0.26}      & \textbf{0.26}                 \\


\bottomrule
\end{tabular}
\caption{ \textbf{Point cloud prediction}. Comparison between \netname and the method of Aubry \etal \cite{aubry14seeing}.}
\label{tab:completion}
\vspace{-2em}
\end{table}

\subsection{Depth prediction}\label{s:exp-depth} 

The monocular depth prediction module of \netname is compared against three baselines: \textbf{\netname-Rand} uses \netname to estimate depth but predicts random confidence scores. \textbf{\netnamedepth} is a variant of the state-of-the-art depth prediction network from \cite{laina2016deeper} based on the same $\PhiDepth$ subnetwork as \netname (but dropping $\PhiPCL$ and $\PhiVP$). Following \cite{laina2016deeper}, for training it uses the BerHu depth loss and a dropout layer, which allows it to produce a confidence score of the depth measurements at test time using the sampling technique of~\cite{kendall2015bayesian,gal2016Bayesian}. Finally, \textbf{\netnamedepth-Rand} is the same network, but predicting random confidence scores.

\myparagraph{Results} Fig.~\ref{fig:depthest} (right) shows the cumulative root-mean-squared (RMS) depth reconstruction error for LDOS after sorting pixels by their confidence as estimated by the network. By fitting better to inlier pixels and giving up on outliers, \netname produces a much better estimate than alternatives for the vast majority of pixels. Furthermore, accuracy is well predicted by the confidence scores. Fig.~\ref{fig:depthest} (left) shows the cumulative RMS by depth, demonstrating that accuracy is better for pixels closer to the camera, which are more likely to be labeled with correct depth. Qualitative results are shown in \cref{fig:depth_qual}. 

\subsection{Point cloud prediction}\label{s:exp-pcc} 

We evaluate the point cloud completion module of \netname by comparing ground truth point clouds $C$ to the point clouds $\hat C$ predicted by $\PhiPCL$ using: (1) the voxel intersection-over-union (VIoU) measure that computes the Jaccard similarity between the volumetric representations of $\hat C$ and $C$, and (2) the normalized point cloud distance of~\cite{rock2015completing}. We average these measures over the test set leading to mVIoU and m$D_{pcl}$ (see supp. material for details).  

\netname is compared against the approach of Aubry~\etal~\cite{aubry14seeing} using their code. \cite{aubry14seeing} 
is a 3D CAD model retrieval method which first trains a large number of exemplar models which, in our case, are represented by individual video frames with their corresponding ground truth 3D point clouds.
Then, given a testing image, \cite{aubry14seeing} detects the object instance 
and retrieves the best matching model from the database. We align the retrieved point cloud to the object location in the testing image using the P3P algorithm. 
For \netname, we evaluate two flavors. The original \netname that predicts the point cloud $\hat C$ and \netname-Fuse which further merges $\hat C$ with the predicted partial depth map point cloud $\hat P$.

\Cref{tab:completion} shows that our reconstructions are significantly better on both metrics for both LDOS chairs and FrC cars. Fusing the results with the original depth map produces a denser point cloud estimate and marginally improves the results. 
Qualitative results are shown in \cref{fig:pcl_qual}.
\section{Conclusion}\label{s:conclusions}

We have demonstrated the power of motion cues in replacing manual annotations and synthetic data in learning 3D object categories. We have done so by proposing a single neural network that simultaneously performs monocular viewpoint estimation, depth estimation, and shape reconstruction. This network is based on two innovations, a new image-based viewpoint factorization method and a new probabilistic shape representation. The contribution of each component was assessed against suitable baselines.

\paragraph*{Acknowledgments.}

We are grateful for support by NAVER LABS Europe and ERC StG 638009-IDIU.

\appendix

\renewcommand\thesection{\Alph{section}}
\renewcommand\thesubsection{\thesection.\arabic{subsection}}
\renewcommand\thefigure{\Alph{figure}}
\setcounter{table}{0}
\renewcommand{\thetable}{\Alph{table}}

\begin{figure*}[ht!]
\centering \Large \bf Learning 3D Object Categories by Looking Around Them \\ \vspace{0.3cm} \normalfont \textit{Appendix}
\end{figure*}

\section{Method: additional details}

\subsection{Scale ambiguity in SFM}\label{s:sfm}

In Sec. 3.2 in the paper, we explain that the scale ambiguity of structure from motion (SFM) causes each reconstruction of a sequence $S^i$ to be known only up to
a global sequence specific scaling factor $\lambda^i$.
Since $\lambda^i$ is not required to learn \PhiVP, but it is important for depth prediction (as discussed in Sec. 3.3 from the paper), we estimate it as well. 

To do so, we note that, given a pair of frames $(t,t^\prime)$ from sequence $S^i$, one can estimate the sequence scale as
$
\lambda^i_{t,t^\prime}
=
\frac
{\| T^i_{t^\prime} - R^i_{t^\prime t} T^i_{t} \|}
{\| \hat T^i_{t^\prime} - R^i_{t^\prime t} \hat T^i_{t} \|}.
$
This expression allows us to conveniently estimate $\lambda^i$ on the fly as a moving average during the SGD iterations used to learn \PhiVP, as samples $\lambda^i_{t,t^\prime}$ can be computed essentially for free during this process.

\subsection{The \netname architecture: further details}\label{s:arch}

\begin{figure*}
\centering
\includegraphics[width=\linewidth]{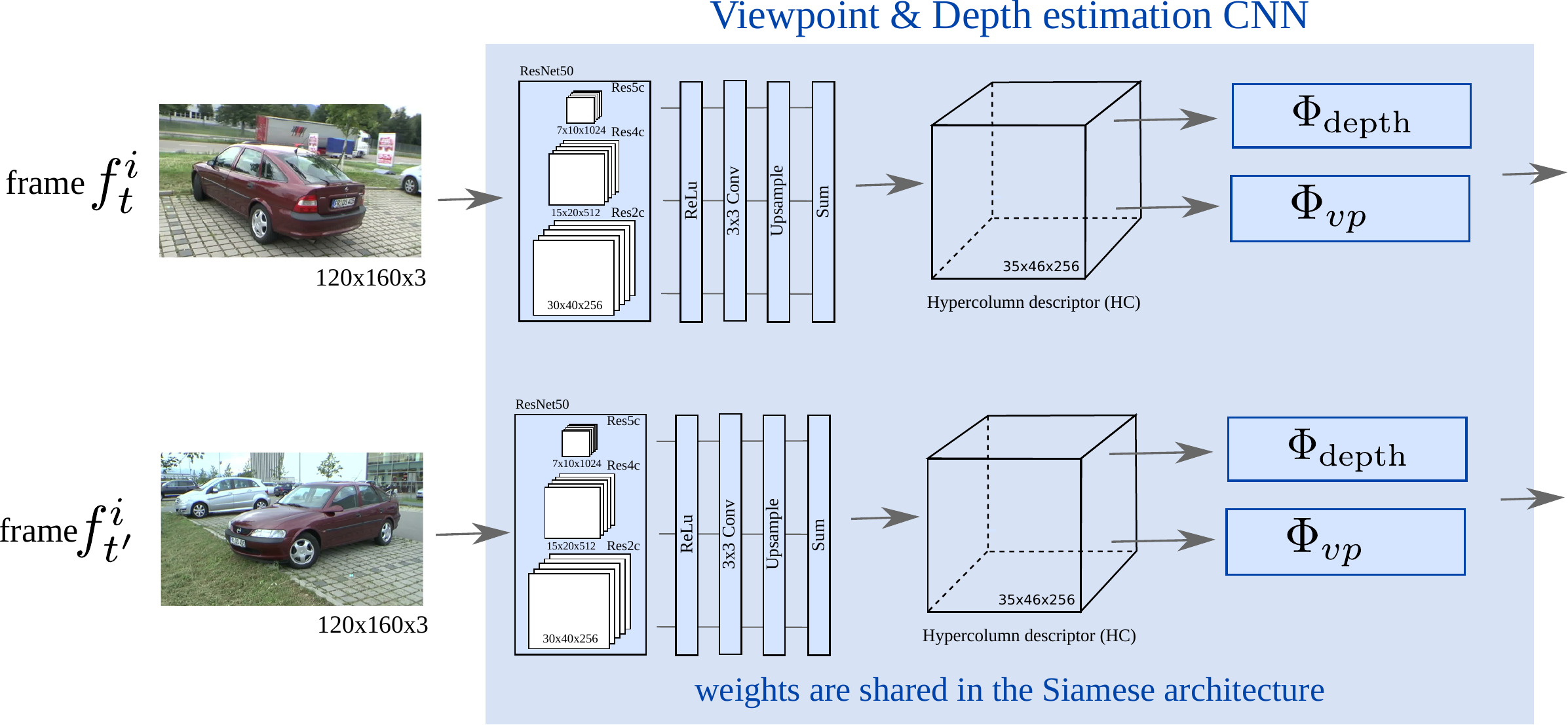}
\caption{\textbf{The core architecture of \netname.} This figure extends the Viewpoint \& Depth estimation block from Figure 2 in the paper and describes the architecture of the hypercolumn (HC) module. \label{fig:overview-hc}}
\end{figure*}

This section contains additional details about the layers that compose the \netname architecture.

\myparagraph{The core architecture}
The architecture of the \netname (introduced in Sec. 3.2 from the paper) is a variant of the ResNet-50 architecture~\cite{he16resnet} with some modifications to improve its performance as a viewpoint and depth predictor that we detail below.

In order to decrease the degree of geometrical invariance of the network, we first replace all $1\times 1$ downsampling filters with full $2\times 2$ convolutions. We then attach bilinear upsampling layers that first resize features from 3 different layers of the architecture (res2d, res3d, res4d) into fixed-size tensors and then sum them in order to create a multiscale intermediate image representation which resembles hypercolumns (HC) \cite{hariharan2015hypercolumns}. An extension of Fig. 2 from the paper that contains the diagram of this HC module can be found in \Cref{fig:overview-hc}. 

\myparagraph{Architecture of the viewpoint factorization network \PhiVP}
HC is followed by 3 modified $3\times 3$ downsampling residual layers that produce the final viewpoint prediction. While the standard downsampling residual layers do not contain the residual skip connection due to different sizes of the input and output tensors, here we retain the skip connection by performing $3\times 3$ average pooling over the input tensor and summing the result with the result of the second $3\times 3$ downsampling convolution branch. We further remove the ReLU after the final residual summation layer. \Cref{fig:overview-vp} contains an overview of the viewpoint estimation module together with a detailed illustration of the modified downsampling residual blocks.

\begin{figure}
\centering
\includegraphics[width=\linewidth]{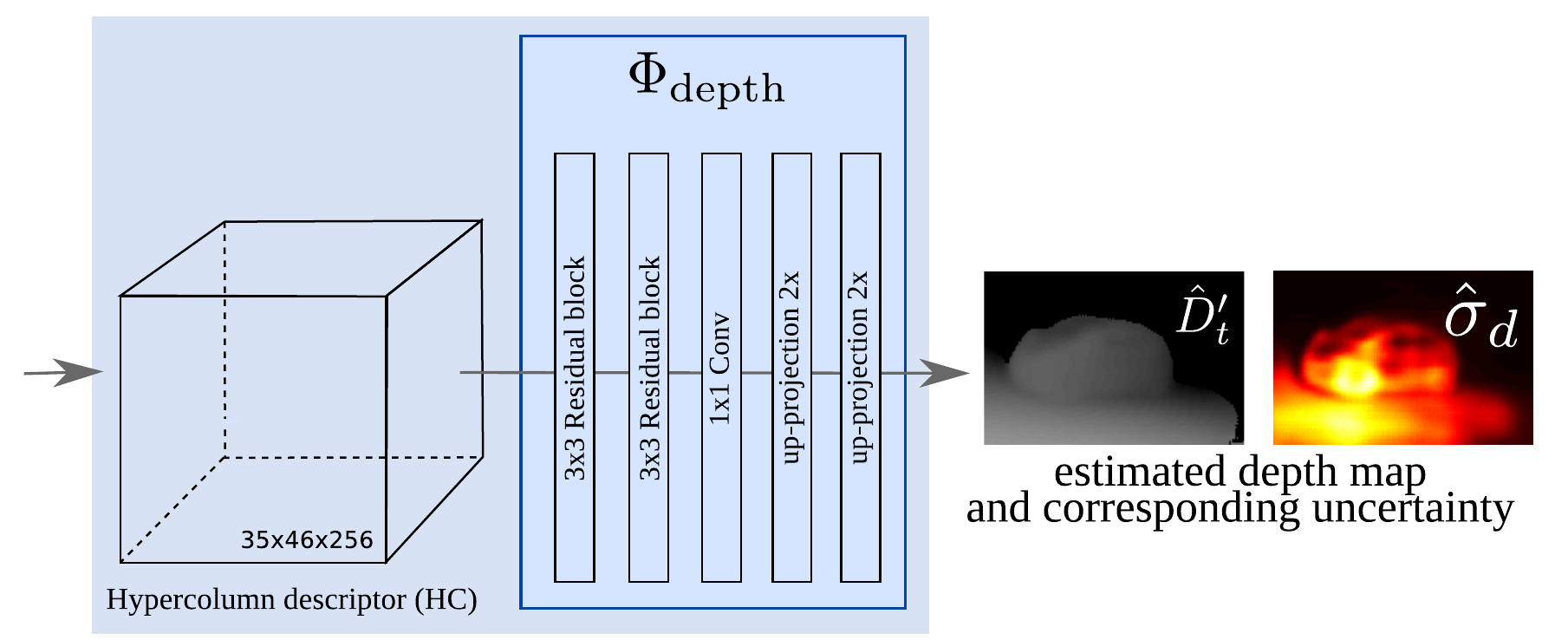}
\caption{\textbf{The architecture of \PhiDepth.} \label{fig:overview-depth}}
\end{figure}

\begin{figure}
\centering
\includegraphics[width=\linewidth]{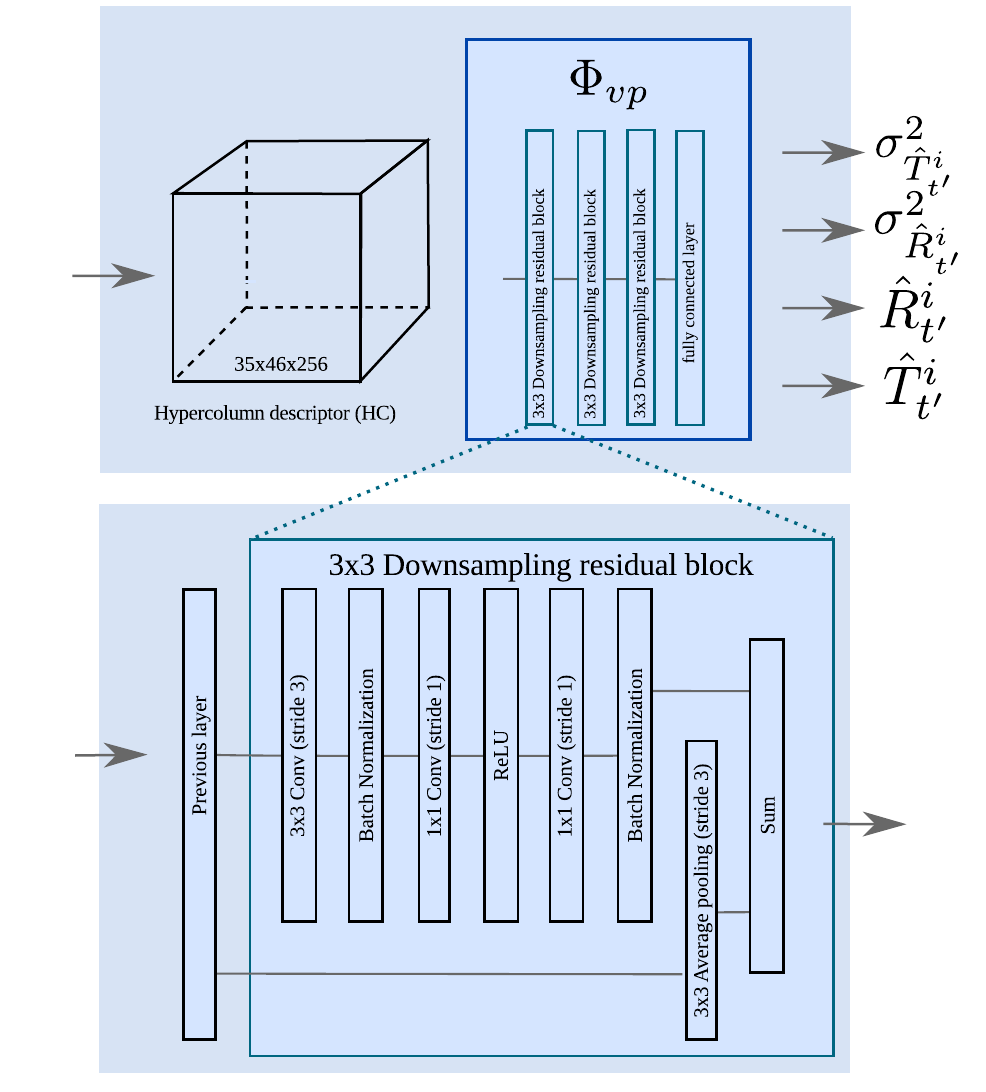}
\caption{\textbf{The architecture of \PhiVP.} Top: the layers of \PhiVP, bottom: A detail of the 3x3 downsampling residual block.\label{fig:overview-vp}}
\end{figure}

\myparagraph{Architecture of the depth prediction  \PhiDepth}
The depth prediction network (introduced in Sec. 3.3 from the paper) shares the early HC layers with the viewpoint factorization network \PhiVP. The remainder of the pipeline is based on the state-of-the-art depth estimation method of \cite{laina2016deeper}. More precisely, after attaching 2 standard residual blocks to the HC layers, the network also contains
two 2x2 up-projection layers from \cite{laina2016deeper} leading to a 64-dimensional representation of the same size as the input image. This is followed by 1x1 convolutional filters that predict the depth and confidence maps $\hat D_t$ and $\hat \sigma_{d_j}$ respectively.
\Cref{fig:overview-depth} contains an illustration of \PhiDepth.

\myparagraph{Architecture of the point cloud completion network \PhiPCL}
Differently from the two previous networks, the point cloud completion network $\PhiPCL$ (introduced in Sec. 3.4 from the paper) is not convolutional but uses a residual multi-layer perceptron (MLP), \ie a sequence of residual fully connected layers.

In more details, the network starts by appending to each 3D point $\hat p_i \in \hat P^G_f \subset \mathbb{R}^3$ an appearance descriptor $a_i$ and processes 
this input with an MLP with an intermediate pooling operator:
$$
(\hat S, \hat \delta) =
\PhiPCL(\hat P^G_f) =
  \operatorname{MLP}_2
  \left(
  \operatornamewithlimits{pool}_{1\leq i \leq |\hat P^G_f|}   
  \operatorname{MLP}_1(\hat p_i, a_i)
  \right).
$$
The intermediate pooling operator, which is permutation invariant, removes the dependency on the number and order of input points $\hat P_f^G$. In practice, the pooling operator uses both max and sum pooling, stacking the results of the two. 

\begin{figure}
\centering
\includegraphics[width=0.9\linewidth]{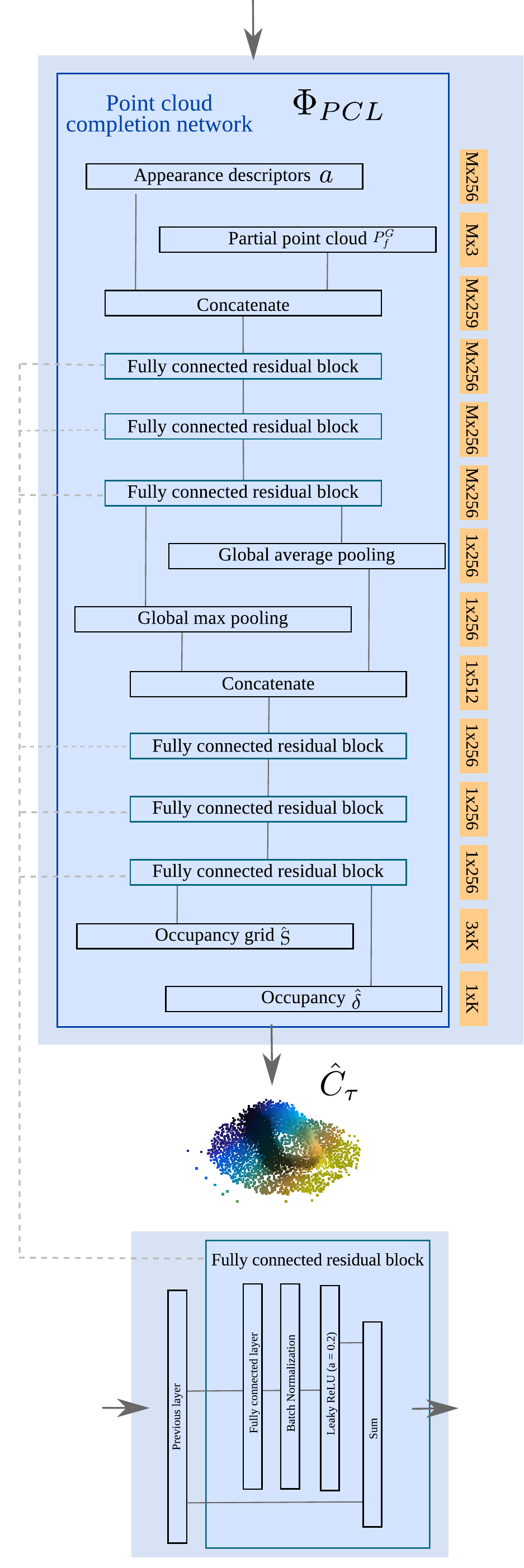}
\caption{\textbf{The architecture of \PhiPCL.} Top: The overview of the point cloud completion network, bottom: A detail of the fully connected residual block.
Orange boxes denote the sizes of the layer outputs. \label{fig:overview-pcl}}
\end{figure}

For the appearance descriptors, recall that each point $\hat p_i$ is the back-projection of a certain pixel $(u_i,v_i)$ in image $f$. To obtain the appearance descriptor $a_i$ we reuse the HC features from the core architecture and sample a column of feature channels at location $(u_i,v_i)$ using differentiable bilinear sampling. Note that, following \cite{tatarchenko16multi}, the fully connected residual blocks contain leaky-ReLUs with the leak factor set to 0.2. A diagram depicting \PhiPCL{} can be found in \Cref{fig:overview-pcl}. 

\section{Experimental evaluation}

In this section we provide additional details about the learning procedures of the baseline networks 
and about the experimental evaluation.

\subsection{Learning details of \netnamedepth and VPNet}
In this section we provide learning details for the BerHu-Net and VPNet baselines.
The learning rates and batch sizes were in all cases adjusted empirically such that the convergence is achieved on the
respective training sets.

\textbf{\netnamedepth} is trained with stochastic gradient descent with a momentum of 0.0005, initial learning rate $10^{-3}$ and a batch size of 16.
The learning rate was lowered tenfold when no further improvement
in the training losses was observed. The BerHu loss
uses the adaptive adjustment of the loss cut-off threshold as explained in \cite{laina2016deeper}.
For the 2x2 up-projection layers we used the implementation of \cite{laina2016deeper}.
For each test image, we repeat the depth map extraction 70 times\footnote{We empirically verified that 70 repetitions are enough for convergence of the variance estimates.}
with the dropout layer turned on and compute the variance of the predictions in order to obtain the per-pixel depth confidence values.
The final feed-forward pass turns off the dropout layer and produces the actual depth predictions.

\textbf{VPNet} is trained with stochastic gradient descent with a momentum of 0.0005, initial learning rate $10^{-2}$ and a batch size of 128. 
The learning rate was lowered tenfold when no further improvement
in the training losses was observed. 
For VPNet trained on aligned FrC, we adjusted the produced bounding box and viewpoint annotations in the same fashion as
done for adjusting the Pascal3D annotations in sec. 5.1. in the paper, 
ensuring that the aligned FrC dataset is as compatible as possible with the target Pascal3D dataset. 
For LDOS, the produced dataset was adjusted in the same way except that we did not use the bounding boxes predicted by \cite{sedaghat15unsupervised}
because the input video frames already focus on full/truncated views of the object category.

\subsection{Additional results}
In sec. 5.1. in the paper we compared \netname to
\cite{sedaghat15unsupervised} on an adjusted version of the Pascal3D dataset. In this section, we additionally report the standard
AVP measure \cite{xiang2014beyond} on the original Pascal3D dataset in order to present a better comparison
with fully supervised state-of-the-art on this dataset. Because the AVP measure requires an object detector,
we extract viewpoints from the same set of RCNN detections as in \cite{tulsiani2015viewpoints}.
Due to the fact that the AVP measure, as well as most other measures from sec. 5.1. in the paper, depends on the dataset-specific 
global alignment transformation $\mathcal{T}_G$, we estimate it from the ground truth annotations
of the training set of \cite{xiang2014beyond} using the same method as described in sec. 5.1. in the paper.

Due to the additional measurement noise brought by the estimation of $\mathcal{T}_G$, we report results only
for the coarsest resolution of 4 azimuth bins. Our \netname obtained 33.4 and 14.7 AVP for the car and chair classes 
vs. 29.4 and 14.3 AVP obtained by \cite{sedaghat15unsupervised} using the same detections from \cite{tulsiani2015viewpoints}. 
Our approach performs on par with some fully supervised approaches such as 3D DPM \cite{pepik2012teaching}, 
while being inferior to the fully supervised state-of-the-art by the same margin as for the other metrics reported in table 1 in the paper.

\subsection{Absolute pose evaluation protocol}
As noted in the paper, the absolute pose error metrics $e_R$ and $e_C$
can be computed only after aligning the implicit global coordinate frames of the benchmarked network
and of the ground truth annotations. This procedure is explained in detail below.

Given a set of ground truth camera poses $g_i^\ast\ = (R_i^\ast,T_i^\ast)$
and the corresponding predictions  $\hat g_i = ( \hat R_i,\hat T_i)$,
we want to estimate a global similarity transform $\mathcal{T}_G = (R_G,T_G,s_G)$, parametrized by a
scale $s_G \in \mathbb{R}$, translation $T_G \in \mathbb{R}^3$ and rotation $R_G \in SO(3)$, such that
the coordinate frames of $g_i^\ast$ and $\hat g_i$ become aligned.

In more detail, the desired global similarity transform
satisfies the following equation:
\begin{equation}
\hat R_i ( R_G X + T_G ) + s_G \hat T_i = R_i^\ast X + T_i^\ast ~ ; ~ \forall X
\label{eq:globadjust}
\end{equation}
\ie given an arbitrary world-coordinate point $X \in \mathbb{R}^3$, its projection into the coordinate frame of $g_i^\ast$ (the right part of \cref{eq:globadjust})
should be equal to the projection of $X$ into the coordinate frame of $\hat g_i$ after transforming $X$ with $R_G$, $T_G$ and 
scaling the corresponding camera translation vector $\hat T_i$ with $s_G$ (the left side of \cref{eq:globadjust}). 
Note that for LDOS data $\mathcal{T}_G$ corresponds to a rigid motion and $s_G=1$.
Given $\mathcal{T}_G$, the adjusted camera matrices $\hat g_i^{ADJUST}$ for which $\hat g_i^{ADJUST} \approx g_i^\ast$ 
are then computed with
$$
\hat g_i^{ADJUST} = ( ~~ \hat R_i R_G ~~ ,  ~~\hat R_i T_G + s_G \hat T_i ~~ )
$$

In order to estimate $\mathcal{T}_G$, $X$ is substituted in \cref{eq:globadjust} with $X = C_i^\ast = -{R_i^\ast}^T T_i^\ast$, 
\ie $X$ is set to be the center of the ground truth camera $g_i^\ast$ which is a valid point of the world coordinate frame.
After performing some additional manipulations, we end up with the following constraint:
\begin{equation}
\frac{1}{s_G} R_G C_i^\ast + \frac{1}{s_G} T_G = \hat C_i ~ ; ~ \forall i
\label{eq:umeyama}
\end{equation}
where $\hat C_i = - \hat R_i^T \hat T_i$ is the center of the predicted camera $\hat g_i$.
Given the corresponding camera pairs $ \{ (g_i^\ast,\hat g_i) \}_{i=1}^N $
the constraint in \cref{eq:umeyama} is converted to a least squares minimization problem:
\begin{equation}
\arg\min_{R_G,T_G,s_G} \sum_{i=1}^N || \frac{1}{s_G} R_G C_i + \frac{1}{s_G} T_G - \hat C_i ||^2
\label{eq:umeyama2}
\end{equation}
and solved using the UMEYAMA algorithm \cite{umeyama1991least}. 

For Pascal3D we estimate $\mathcal{T}_G$ from the held-out training set
and later use it for evaluation on the test set.
For LDOS, due to the absence of a held-out annotated training set,
we estimate $\mathcal{T}_G$ on the test set.

\subsection{Point cloud prediction}
The normalized point cloud distance of~\cite{rock2015completing} is computed as
$
D_\text{pcl}(C,\hat C) = 
\frac{1}{|C|}
\sum_{c \in C}
\min_{\hat c \in \hat C}
\| \hat c - c \| 
+ 
\frac{1}{|\hat C|}
\sum_{\hat c \in \hat C}
\min_{c \in C} \|\hat c - c\|.
$
For the VIoU measure, a voxel grid is setup around each ground truth point-cloud $C$ by uniformly subdividing $C$'s  bounding volume into $30^3$ voxels. 

The point clouds are compared within the local coordinate frames of each frame's camera (whose focal length is assumed to be known). Furthermore, since the
SFM reconstructions are known only up to a global scaling factor, we adjust each point cloud prediction $\hat C$ from the FrC dataset by multiplying it with
a scaling factor $\zeta$ that aligns the means of $\hat C$ and $C$. Note that $\zeta$ can be computed analytically with:
$$
\zeta = \frac{\mu_{C}^T \mu_{\hat C} } { \mu_{\hat C}^T \mu_{\hat C} },
$$
where $\mu_{C} = \frac{1}{|C|} \sum_{c_m \in C} c_m $ is the centroid of the point cloud $C$.

\myparagraph{Ablative study} In table 2 in the paper, we have presented a comparison of \netname to the baseline approach from \cite{aubry14seeing}.
Here we provide an additional ablative study that evaluates the contribution of the components of $\PhiPCL$.
More exactly, \cref{tab:completion} extends table 2 from the paper with the following flavours of \netname:
(1) \netname-$\hat P_f$ which only predicts the partial point cloud $P_f$,
(2) \netname-Chamfer which removes the density predictions $\hat \delta$ and replaces $l_{pcl}(\hat S)$ with a Chamfer distance loss and
(3) \netname-$\hat S$ that predicts the raw unfiltered and untruncated point cloud $\hat S$. 

The drops in performance by predicting solely the raw and partial point clouds $\hat P_f$ and $\hat S$ emphasize the importance of the point cloud completion and density prediction components respectively. The Chamfer distance loss brings marginal improvements in $D_{pcl}$ but a significant decrease of VIoU due to the inability of the network to represent and discard outliers.

\begin{table}
\centering 
\footnotesize
\setlength\tabcolsep{2pt}
\begin{tabular}{lrrrr}
\toprule
Test set                          & \multicolumn{2}{c}{\textbf{LDOS}} & \multicolumn{2}{c}{\textbf{FrC}} \\
\midrule
Metric                            &  $\uparrow$ mVIoU    & $\downarrow$ m$D_{pcl}$  & $\uparrow$ mVIoU     & $\downarrow$ m$D_{pcl}$   \\
\midrule
Aubry \cite{aubry14seeing}       & 0.06      & 1.30        & 0.21      & 0.41                 \\
\netname-$\hat P_f$     & 0.10      & 0.37        & 0.11      & 0.56 \\
\netname-Chamfer        & 0.09	     & \textbf{0.18}	       & 0.20	   &\textbf{0.24} \\
\netname-$\hat S$       & 0.12      & 0.27        & 0.18      & 0.50                 \\
\textbf{\netname (ours) }         & \textbf{0.13}      & 0.20        & 0.24      & 0.28                 \\
\textbf{\netname-Fuse (ours)}    & \textbf{0.13}      & 0.19        & \textbf{0.26}      & 0.26                 \\
\bottomrule
\end{tabular}
\caption{ \textbf{Point cloud prediction ablative study}. 
Comparison between \netname and the method of Aubry \etal \cite{aubry14seeing} and an additional ablative study.}
\label{tab:completion}
\vspace{-2em} 
\end{table}

\myparagraph{Related methods} Note that apart from \cite{aubry14seeing}, there exist newer works that tackle the problem of single-view
3D reconstruction \cite{huang2015single,massa2016deep}, however these were not considered due to their requirement of renderable
mesh models which are not available in our supervision setting.

\clearpage

{\small\bibliographystyle{ieee}\bibliography{refsclean}}

\end{document}